\documentclass[Afour, sageh,times]{sagej}

\usepackage{moreverb,url}

\usepackage[colorlinks,bookmarksopen,bookmarksnumbered,citecolor=red,urlcolor=red]{hyperref}
\usepackage{lineno,hyperref}
\usepackage{color,soul}
\usepackage{bm}
\usepackage{amssymb}
\usepackage{amsmath}
\usepackage{float}
\usepackage{times}
\usepackage{balance}
\usepackage{breqn}
\usepackage{threeparttable}
\usepackage{multirow}
\usepackage{booktabs}
\usepackage{makecell}
\usepackage{bigstrut}
\usepackage{graphics} 
\usepackage{graphicx}
\usepackage{epsfig} 
\usepackage{wrapfig}
\usepackage[table]{xcolor}

\newcommand\BibTeX{{\rmfamily B\kern-.05em \textsc{i\kern-.025em b}\kern-.08em
T\kern-.1667em\lower.7ex\hbox{E}\kern-.125emX}}

\def\volumeyear{2024}

\begin{document}

\runninghead{Xing and Buschka}

\title{Understanding Human Activity with Uncertainty Measure for Novelty in Graph Convolutional Networks}

\author{Hao Xing\affilnum{1} and Darius Burschka\affilnum{1}}

\affiliation{\affilnum{1}Authors are with Machine Vision and Perception Group, School of Computation, Information and Technology, Technical University of Munich, Germany
}

\corrauth{Hao Xing, Machine Vision and Perception Group, School of Computation, Information and Technology, Technical University of Munich, Boltzmannstraße 3, 85748 Garching, Germany.}
\email{hao.xing@tum.de}

\begin{abstract}
{Understanding human activity is a crucial aspect of developing intelligent robots, particularly in the domain of human-robot collaboration. Nevertheless, existing systems encounter challenges such as over-segmentation, attributed to errors in the up-sampling process of the decoder. In response, we introduce a promising solution: the Temporal Fusion Graph Convolutional Network. This innovative approach aims to rectify the inadequate boundary estimation of individual actions within an activity stream and mitigate the issue of over-segmentation in the temporal dimension.

Moreover, systems leveraging human activity recognition frameworks for decision-making necessitate more than just the identification of actions. They require a confidence value indicative of the certainty regarding the correspondence between observations and training examples. This is crucial to prevent overly confident responses to unforeseen scenarios that were not part of the training data and may have resulted in mismatches due to weak similarity measures within the system. To address this, we propose the incorporation of a Spectral Normalized Residual connection aimed at enhancing efficient estimation of novelty in observations. This innovative approach ensures the preservation of input distance within the feature space by imposing constraints on the maximum gradients of weight updates. By limiting these gradients, we promote a more robust handling of novel situations, thereby mitigating the risks associated with overconfidence. Our methodology involves the use of a Gaussian process to quantify the distance in feature space.}



The final model is evaluated on two challenging public datasets in the field of human-object interaction recognition, i.e., Bimanual Actions and IKEA Assembly datasets, and outperforms popular existing methods in terms of action recognition and segmentation accuracy as well as out-of-distribution detection.
\end{abstract}


\maketitle

\section{Introduction}
\label{sec:1}

Understanding Human-Object Interactions plays an important role in intelligent systems, especially for robots to learn from demonstrations and collaborate with humans, which requires not only recognizing and segmenting interaction relations per frame but also quantifying prediction uncertainty.

The quality of an action recognition system relies on the identification of cues that define the label of an action, as well as the spatial-temporal relations that define the boundaries between consecutive actions in a task. In our previous work~\cite{xing2022understanding}, we formulated representations for activities within spatio-temporal graphs. These graphs feature human skeleton joints and the central points of bounding boxes encapsulating objects as graph nodes, where graph edges signify the dynamic relationships between these nodes. Sub-activities are recognized and segmented frame-wise by analyzing the dynamic connections between graph nodes. In that work, we have proven that the attention-based Graph Convolutional Network (GCN) is one of the most promising solutions to process dynamic HOI graph relations. It adaptively updates the correlations between nodes through an attention mechanism, and iteratively parses features in spatial and temporal dimensions. In conjunction with a \textit{temporal pyramid pooling} (TPP) decoder, the processed graph features underwent further upsampling to revert to the original temporal scale. Subsequently, the resultant graph sequences were classified and segmented frame by frame. However, directly interpolating temporal pooling features diminishes the smoothness and accuracy of segmentation. Overcoming segmentation inaccuracies and over-segmentation in the temporal dimension remains challenging for researchers. Building upon prior successful methodologies, the present work adopts the encoder-decoder framework, and leverages attention-based graph convolutional networks to analyze dynamic graph features, with a specific focus on mitigating the over-segmentation issue.

In order to improve the HOI segmentation performance, we propose a novel Temporal Fusion Graph Convolutional Network (TFGCN), which consists of an attention-based graph convolutional encoder and a newly designed temporal fusion (TF) decoder. The new decoder extracts global features through multiple parallel temporal-pyramid-pooling blocks and enriches temporal features by fusing high-dimensional features from the encoder to processed low-dimensional features. The experimental results on public datasets show better performance in terms of recognition accuracy and preventing boundary shifts and over-segmentation. However, learning-based models are commonly overconfident in wrong predictions, while real scenarios have many unexpected situations, such as noise and unknown data. These factors increase the risk and difficulty of the application. Therefore, the detection of novel human actions is necessary for the implementation of our model.

Multi-object tracking algorithms give inspiration on how to solve the problem, which typically assigns IDs based on the distance between representation features and the existing feature
space~\cite{Wojke2018deep}. In other words, it requires the model to be distance-aware in the representation space~\cite{sngp}, as expressed as follows:

\begin{equation}
\begin{gathered}
    \label{eq:lipschitz}
    \alpha\parallel\bm{x} - \bm{x'}\parallel_{X}< \parallel g(\bm{x}) - g(\bm{x'})\parallel_{G}\ \\ < \  \beta\parallel\bm{x} - \bm{x'}\parallel_{X}
\end{gathered}
\end{equation}
where $g$ means the graph convolutional layer and maps the input data from manifold $X$ (input space) to the representation space $G$ (feature space), $\bm{x}$ and $\bm{x'}$ are two different inputs. The parameters $\alpha$ and $\beta$ are the lower and upper bounds with a constraint of $0<\alpha<\beta$. In this \textit{bi-Lipschitz} condition, the upper bound affects the sensitivity of hidden representations to the novel observations (out-of-distribution, OOD), and the lower bound guarantees the distance in hidden representation space for meaningful changes in the input manifold~\cite{sngp}.  

Traditional cascaded convolutional networks provide an upper bound for the hidden representation space distance through normalization and activation functions~\cite{ruan2018reachability}. However, they suffer from the problem of exploding and vanishing gradients. 

Residual connections show the ability to compensate for the gradient problems ~\cite{veit2016residual}, but lead to a higher bound range and indistinguishable features in the representation space for OOD detection. In order to preserve the meaningful isometric property in our deterministic model, we introduce a Spectral Normalized Residual (SN-Res) connection, which places an upper \textit{Lipschitz} constraint on the residual flow. We form an Uncertainty Quantified Temporal Fusion Graph Convolution Network (UQ-TFGCN) with this novel construction, in which the hidden representation space is restricted to a reasonable region. The final labels of the unknown data, along with their similarity to known data, are predicted using maximum likelihood estimation within a Gaussian Process (GP) kernel.
    
Overall, the technical contributions of the paper are:
\begin{itemize}
	\item we propose a Temporal Fusion Graph Convolution Network (TFGCN) that utilizes a novel temporal feature fusion decoder to enhance the capabilities of Graph Convolutional Networks (GCNs) to understand human-object interaction;
	\item We find that residual connections prove advantageous in maintaining input distances within a familiar space, and we enhance the capability to detect out-of-distribution instances by incorporating spectral normalization of residual connections. We investigate the impact of the spectral normalization coefficient on the accuracy and out-of-distribution detection performance.
	\item we evaluate our model on two challenging, public HOI datasets. Compared to other current action recognition and segmentation approaches, our model achieves the best performance on both datasets in terms of accuracy and uncertainty estimation.
\end{itemize}


\section{Related Work}
\label{sec:2}

Graph convolutional networks have evolved rapidly in recent years in the field of understanding human activities. The following section briefly introduces the most relevant graph convolutional networks,  human-object-interaction understanding methods, and uncertainty quantification methods. Certain sections on graph convolutional networks and human-object interaction understanding are adapted from our prior work~\cite{xing2022understanding}.

\subsection{Graph convolution networks}
Recently, Graph Convolution Networks (GCNs) have been successfully implemented in the field of human action recognition with structured representations.

The GCNs can be categorized into two classes: spatial and spectral. The spatial GCNs operate the graph convolutional kernels directly on spatial graph nodes and their neighborhoods. \cite{yan2018spatial} proposed a Spatial-Temporal Graph Convolutional Network (ST-GCN), which extracts spatial features from the skeleton joints and their naturally connected neighbors and temporal features from the same joints in consecutive frames. \cite{shi2019two} introduced a two-stream adaptive Graph Convolutional Network (2s-AGCN) based on ST-GCN, which adopted an initial attention mechanism in spatial layers to adaptively update the adjacency matrix for multi-input streams (skeleton joints and bones). \cite{chen2021channel} proposed a Channel-wise Topology Refinement Graph Convolution Network (CTR-GCN) that refines a spatial attention mechanism on channel dimension to efficiently learn dynamical features in different channels. \cite{xing2022skeletal} introduced a hybrid attention-based graph convolutional network (HA-GCN), which mixes two attention mechanisms to enrich graph features from different input streams.
{\cite{ling2023sa} introduced a bi-stream (joint and bone) spatial graph convolutional network to detect eye contact for conveying information and intent in wild environments.}

The spectral GCNs consider the graph convolution in the form of spectral analysis \cite{li2015gated}. \cite{henaff2015deep} developed a spectral network incorporating with graph neural network for the general classification task. \cite{kipf2016semi} extends the spectral convolutional network further in the field of semi-supervised learning on graph-structured data.

This work follows the spatial GCNs adopts the aforementioned methods one by one and compares their performance with different encoder-decoder setups.

\subsection{Understanding of human-object-interaction}

The understanding of Human-Object-Interaction (HOI) plays a pivotal role in understanding human activities, which involves the segmentation and recognition of sub-activities framewise through the analysis of interactive relations between human and objects. 

\cite{feichtenhofer2016convolutional} introduced a two-stream 2D CNN that utilizes features from both appearances in still images and stacks of optical flow. \cite{carreira2017quo} proposed a two-stream inflated 3D CNN (I3D) that improves the ability of 2D CNNs in extracting spatial-temporal features. 

With the successful applications of Graph Neural Networks (GCNs) in action recognition, many recent works represent human and objects as nodes in a graph and extract their relation features by GCNs. \cite{datasetKIT} presented a graph network that uses three multi-layer perceptron (MLP) blocks to update nodes, edges, and aggregation features from the graph representation of HOI. The authors also published their HOI dataset, namely the Bimanual Actions dataset. Asynchronous-Sparse Interaction Graph Networks (ASSIGN) \cite{morais2021learning} is another attempt at the HOI recognition and segmentation task. It used a recurrent graph network that automatically detects the structure of interaction events associated with entities of a sequence of interactions, which are defined as human and objects in a scene. More recently, \cite{xing2022understanding} proposed a Pyramid Graph Convolutional Network (PGCN) that adaptively updates the human-object relations by attention mask and upsamples the HOI features to the original time scale by a novel temporal pyramid pooling (TPP) decoder. However, directly interpolating temporal pooling features reduces the smoothness and accuracy of segmentation. \cite{lagamtzis2023exploiting} exploited spatial and temporal hand-object relations by leveraging an encoder-decoder framework with graph neural networks. The network can recognize the hand action label and forecast the next motion by a multilayer perceptron module. However, the authors represented human appearance features by a single graph node, which weakens the performance of action recognition. \cite{tran2023persistent} modeled human-object interactions through a long-term activity route (persistent process) and short-term sub-actions (transient processes). Instead of recognizing action labels, authors focus on the 2D/3D trajectory prediction of the whole activity. Besides single-person action recognition, multi-person involved HOI understanding is another important task. To address the occlusion issue in multi-person actions, \cite{qiao2022geometric} combined both visual and geometry HOI features together and processed features through a Two-level Geometric feature-informed Graph Convolutional Network (2G-GCN). \cite{reily2022real} represented individual actions as graph nodes, interactions between people as graph edges, and finally extracted team intentions from graph features.

Most existing recurrent networks exhibit commendable real-time performance yet are hindered by limited short-term memory. The encoder-decoder structure provides a promising solution for this problem by facilitating a comprehensive fieldview. Compared with multi-person intent recognition, single-person action recognition is a more fundamental and challenging task. Motivated by the successful implementation of the temporal pooling decoder, this study adopts the encoder-decoder structure to enhance the performance of single-performer action understanding. The approach involves extracting global features by the temporal pooling module and fusing condensed features into the temporally pooled features.

\subsection{Uncertainty quantification}
\begin{figure*}[t]
  \centering
    \includegraphics[width=0.9\linewidth]{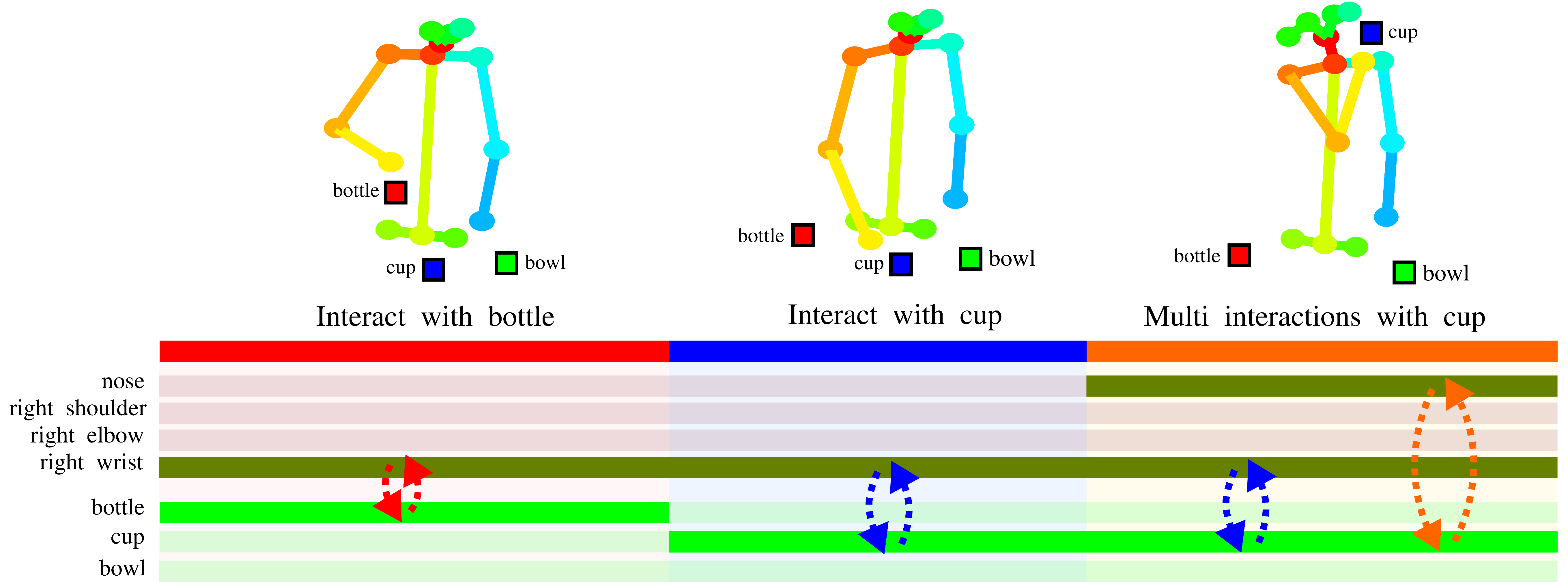}
  \caption{Dynamic relations (depicted by dashed arrows) exist between the body parts and objects involved in the action of \textit{drinking}, with the human represented by a skeleton and objects represented by square boxes. The highlighted time bar signifies active participation in the interaction by either objects or skeleton joints.}
  \label{fig:dynamic_relations}
\end{figure*}

Most existing learning-based classification methods have high accuracy in in-distribution datasets, but suffer from overconfidence in wrong predictions and barely detect out-of-distribution samples.
Many promising works train Bayesian neural networks to approximate uncertainty~\cite{kendall2015bayesian, blundell2015weight}, but it is difficult to converge in a large range of data. \cite{lakshminarayanan2017simple} introduced a method that estimates the prediction uncertainty by ensembling predictions from multiple models. \cite{gal2016dropout} proposed Monte Carlo Dropout (MC-Dropout) to approximate the Bayesian probability. Due to their reliability, these two methods are usually considered as baselines for uncertainty quantification, although they are time-consuming. 

Recently, deterministic network uncertainty quantification (DUQ) methods have been proposed to efficiently estimate the prediction uncertainty in a single forward pass. The key factor is distance awareness in the representation space. \cite{miyato2018spectral} introduced Spectral Normalization to regulate weight update gradients in a generative network. \cite{duq} adopted two sides \textit{Lipschitz} constraints to enforce the gradient smoothness and sensitivity to meaningful changes. Following the two sides \textit{Lipschitz} constraints, \cite{sngp} utilized the spectral normalized kernel instead of normal convolutional kernels on the mainstream to constrain the weight update max gradient and guarantee distance awareness in feature space. However, general Spectral Normalization models exhibit a substantial number of trainable parameters and necessitate considerable computational resources. 

In this study, we observe that the residual connection contributes to maintaining input distance within proximity and has fewer trainable parameters compared to the mainstream. However, it results in an elevation of the \textit{Lipschitz} bounds. To enhance the efficacy of distance awareness in the feature space, we employ Spectral Normalization on the residual connections (SN-Res). Additional empirical evidence supporting this observation is provided in the experiment section.

To measure the uncertainty, the Gaussian Process has been widely implemented. \cite{sngp} approximated the Gaussian Process prior distribution by a learnable neural kernel and further obtained the likelihood per the Laplace approximation. Although it improved the efficiency of the computation of likelihood, but damaged the accuracy of the likelihood. \cite{li2021uncertainty} improved the flash radiography reconstruction by removing the outliers with high uncertainty, which is estimated by the Gaussian probability density function with mean zero and a measured covariance matrix. \cite{su2023uncertainty} utilized a multivariate Gaussian distribution to estimate the uncertainty of each corner of predicted bounding boxes in a LiDAR point cloud, and improved the performance of object detection for autonomous vehicles.

Inspired by these previous works, we follow the uncertainty quantification techniques for deterministic networks and utilize the multivariate Gaussian distribution to model the uncertainty and measure it through its likelihood.



\section{Uncertainty quantified temporal-fusion graph convolutional network}
\label{sec:3}

In this section, we introduce an Uncertainty Quantified Temporal Fusion Graph Convolutional Network (UQ-TFGCN) with Spectral Normalized Residual connection, which balances the distance-preserving ability in representation space and high-accuracy performance.


\subsection{Temporal-fusion graph convolutional network}

The basic idea of temporal fusion graph convolutional network is inspired by image segmentation approaches, which predict the semantic meaning of each pixel unit by extracting global spatial features and mapping it to the corresponding spatial position. In contrast, we feed the graph representations of HOI instead of images into the network, since the graph representation is insensitive to the background and appearance noise. Our temporal fusion graph convolutional network processes the graph features not only in the spatial dimension but also in the temporal dimension. These features are compressed into a low temporal dimensional space by the encoder and upsampled to the original temporal dimension by the decoder.

\subsubsection{Graph construction}

\begin{figure}[t]
    \centering
    \begin{tabular}{@{}cc@{}}
    \includegraphics[width=0.25\textwidth]{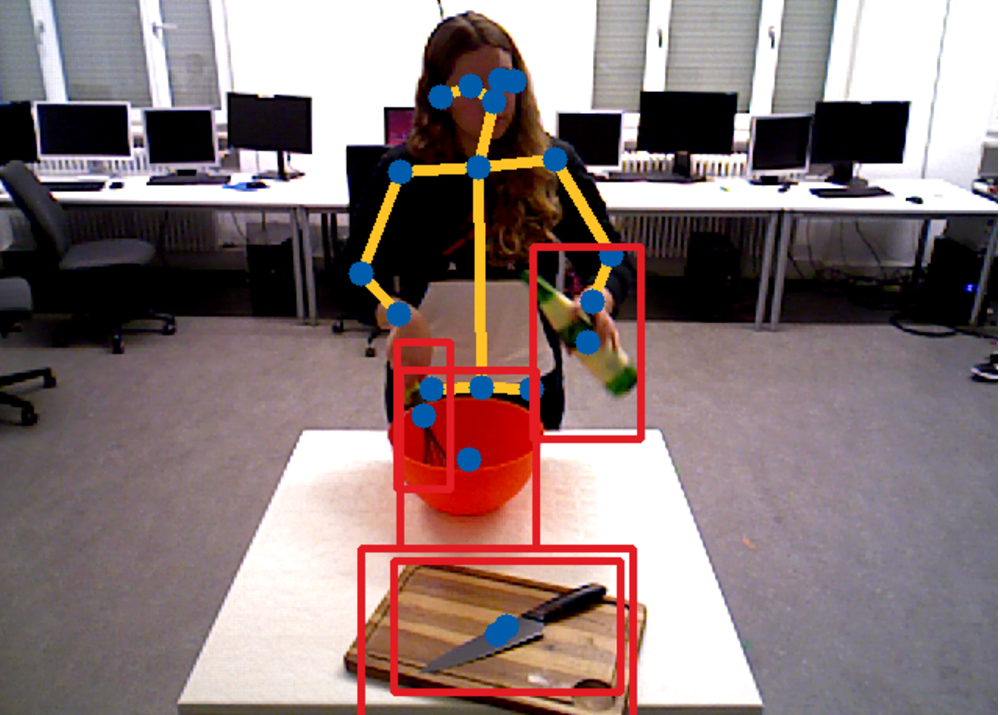} &
    \includegraphics[width=0.19\textwidth]{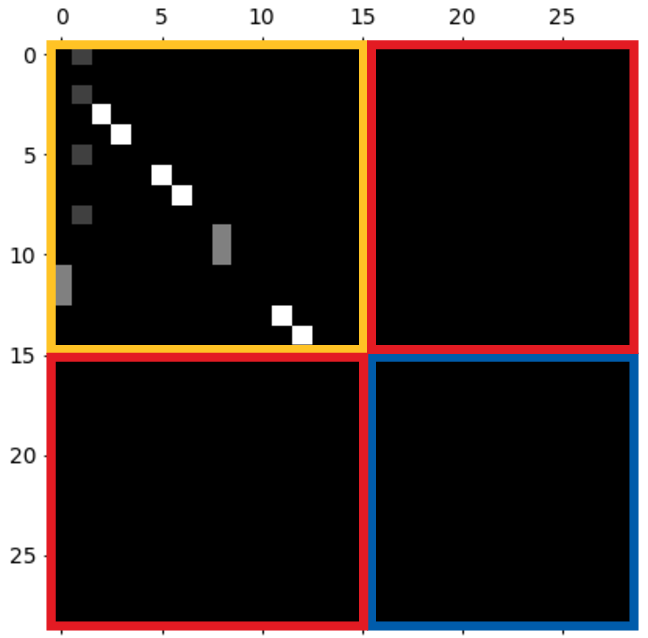} \\
    (a) & (b)
    \end{tabular}
  \caption{\label{fig:graph} Simplify the human-object-interaction into a graph representation from the work~\cite{xing2022understanding}: (a) spatial graph with nodes (blue) and edges (orange) for an example in the Bimanual Actions dataset \cite{datasetKIT}; (b) initial inwards adjacent matrix with skeleton inward edges (orange block), empty human-objects (red blocks) and objects-objects edges (blue block).}
  \label{fig:scene_graph}
\end{figure}

In this work, we focus on the interaction between a single performer and multiple objects. Most of the human body can be conceptualized as an articulated system with joint nodes connected by bones, and most objects can be represented by central points. Therefore, the human-object-interaction (HOI) scenario can be reduced to a graph with nodes and edges, as shown in Fig~\ref{fig:graph} (a). In an HOI graph, each node has three types of edges, namely inward, outward, and self-connecting edges \cite{shi2019two}. The edges of human skeleton are defined by the pose estimation model, which is with inward connections from each joint to adjacent joints that are closer to the center of the body (\textit{neck}), and outward connections in reverse, as shown in Fig~\ref{fig:graph} (b). However, object-related connections (human-objects and objects-objects) are challenging due to the dynamic nature of the scene. As illustrated in Fig~\ref{fig:dynamic_relations}, dynamic relationships are evident in the process of \textit{drinking}. The pertinent targets, including objects and skeleton joints, are highlighted along the time axis to depict sub-actions. Specifically, the \textit{right wrist} node sequentially interacts with the \textit{bottle} and \textit{cup} node. In the final phase of the action, the \textit{cup} node concurrently engages with both the \textit{right wrist} and \textit{nose}. Traditional spatial graphs are predefined based on prior knowledge, which barely accommodates these dynamic relationships. Hence, in this work, we employ a spatial attention mechanism in the encoder to adaptively update dynamic edges.

\subsubsection{Encoder}
\begin{figure*}[t]
  \centering
    \includegraphics[width=0.90\linewidth]{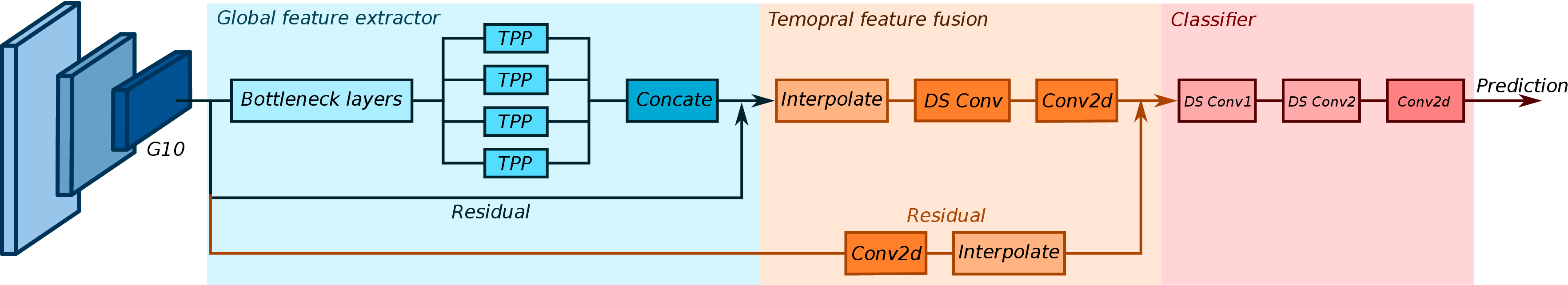}
  \caption{Framework of temporal fusion decoder including three blocks: temporal feature extractor, feature fusion, and classifier. The ”Concate” block concatenates all feature maps from \textit{temporal pyramid pooling} (TPP) layers into one. ”DS Conv” represents a depth-wise 2D convolutional layer.}
  \label{fig:decoder}
\end{figure*}

\begin{figure}[t]
  \centering
    \begin{tabular}{@{}cc@{}}
    \includegraphics[width=0.5\linewidth]{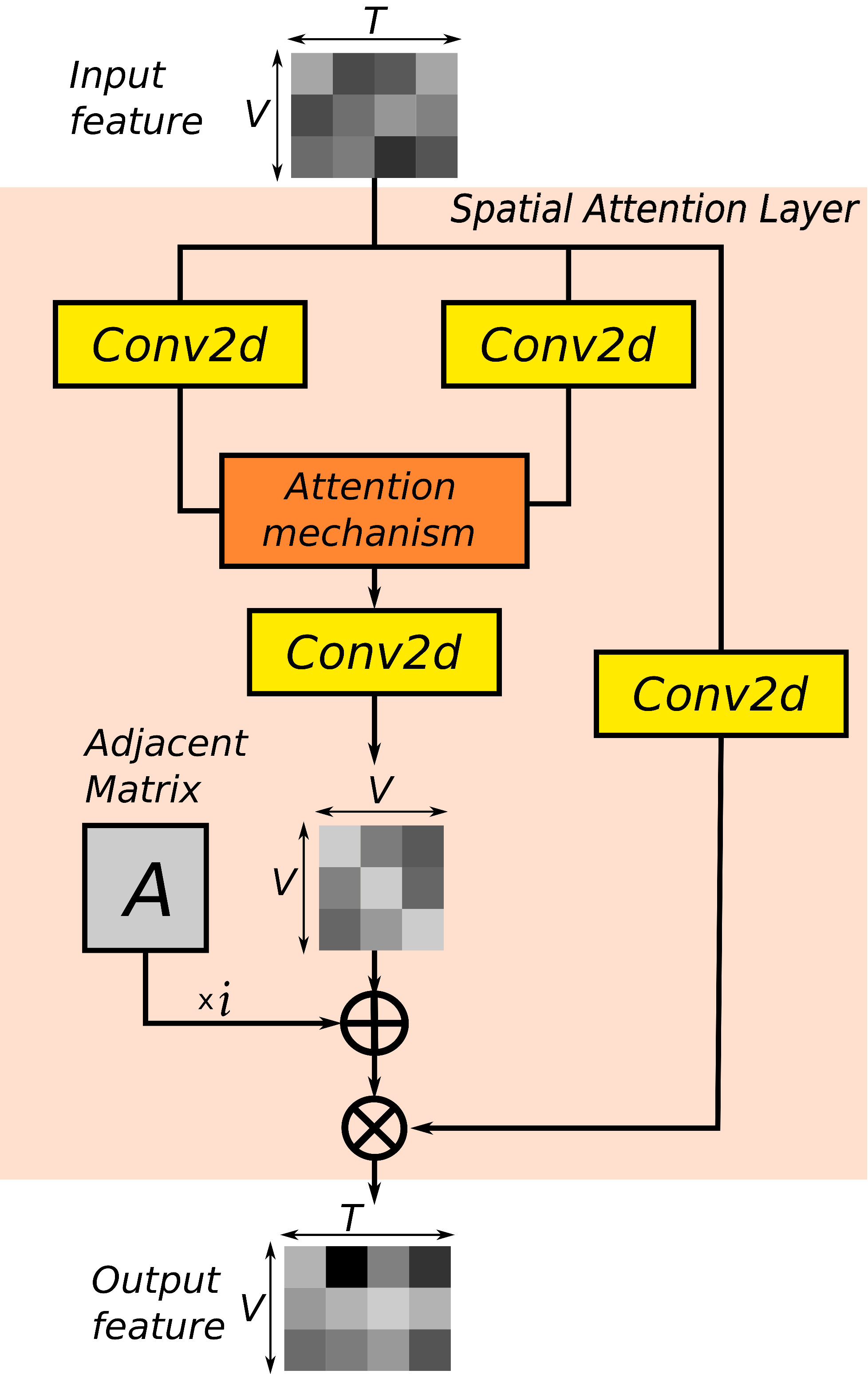} &\ \ \ \ 
    \includegraphics[width=0.26\linewidth]{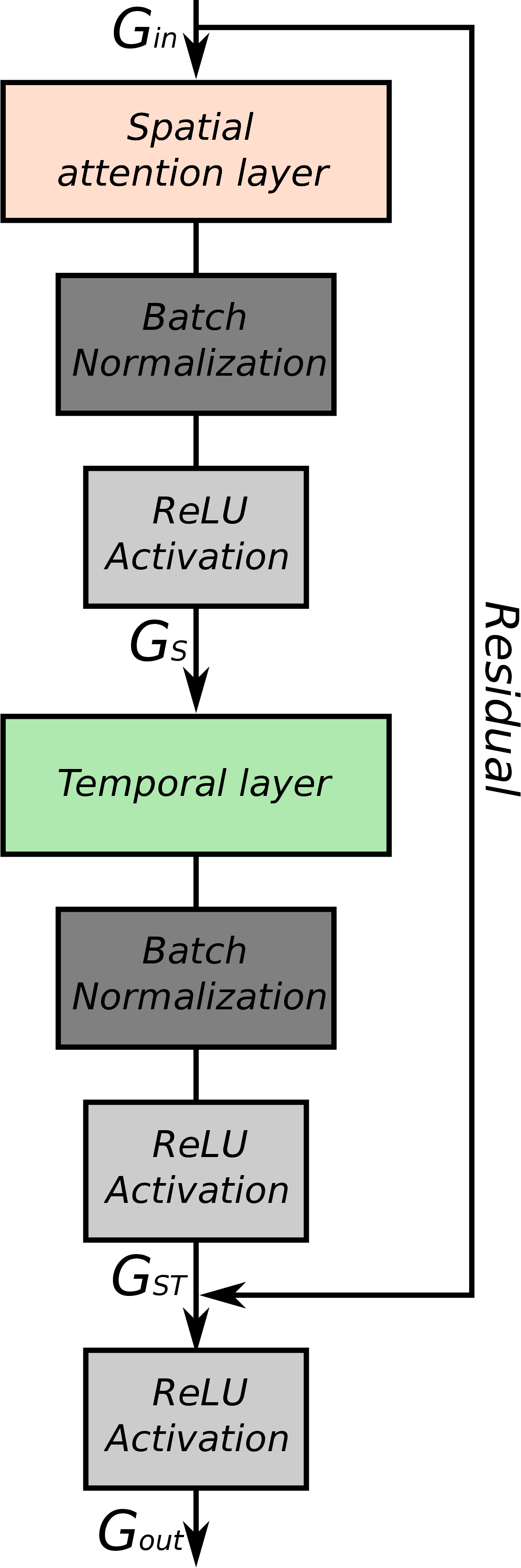}\\
    (a) & (b)
    \end{tabular}
  \caption{Structure of the Spatial-Temporal Graph Convolutional block. (a) A spatial graph convolutional layer with the attention map $\bm{A}_{att}$ and a trainable parameter $a$ for adaptive update of the adjacency matrix $\bm{A}$. (b) A basic block unit consisting of the spatial convolutional layer (see a), Batch Normalization, temporal layer, and ReLU activation function with a residual side branch.}
  \label{fig:attention_spatial}
\end{figure}

As shown in Fig~\ref{fig:attention_spatial} (a), the core theory of attention mechanism is updating the predefined adjacent matrix by adding global correlations \textemdash\ the attention map that can generally be calculated in the following way:
\begin{equation}
    a_{ij} = \phi(\frac{{\bm{f}_i}^T \cdot \bm{f}_j}{\sqrt{n}}) \ \ \text{or}\  \phi(\Bar{f}_i-\bar{f}_j)
\end{equation}
where $\phi$ is an activation function, e.g., \textit{hyperbolic tangent} function, $a_{ij}$ is an element of the attention map $\bm{A}_{tt}$, and $i$, $j$ are its index. $\bm{f}$ represents a vector of the feature map, $\bar{f}$ means the average value of the feature vector, and $n$ serves as a normalization factor, typically selected to be the length of the feature vector.

Given an input graph $\bm{G}_{in} \in \mathbb{R}^{C_{in}\times T\times V}$, the spatial graph feature map $\bm{G}_s\in \mathbb{R}^{C_{out}\times T\times V}$ can be obtained by the following equation:
\begin{equation}
\begin{gathered}
    \bm{G}_s =  Conv2d(\bm{G}_{in}) \cdot (a\bm{A}_{att} + \bm{A}),\\ \bm{A}_{att}, \bm{A}\in\mathbb{R}^{V\times V}
\end{gathered}
\end{equation}
where $C_{in}$, $C_{out}$, $T$, and $V$ are the input channel number, output channel number, temporal size, and spatial size, respectively. Note that the kennel size of the convolutional kernel in the spatial layer is $1 \times 1$, since the spatial features are processed by the attention map and adjacent matrix. 

The spatial graph feature is further processed by a temporal graph convolutional layer to obtain the spatio-temporal processed feature map $\bm{G}_{st}$, as shown in Fig~\ref{fig:attention_spatial} (b). A simple example is the temporal layer used in ST-GCN~\cite{yan2018spatial}, which is a single 2D convolutional kernel with kernel size $9$ in temporal dimension. The final output graph feature map $\bm{G}_{out}$ is then obtained by merging spatial-temporal processed feature map $\bm{G}_{st}$ with a residual connection as follows:
\begin{equation}
    \bm{G}_{out} =  res(\bm{G}_{in})+ \bm{G}_{st}
\end{equation}

To achieve the best performance, we compare the effect using several popular existing attention-based graph convolutional networks as encoders. The ST-GCN~\cite{yan2018spatial} is implemented as a baseline without an attention mechanism. The AGCN~\cite{shi2019two} is a variant of ST-GCN that adds a product attention mechanism to the spatial graph convolution layer. The encoder of PGCN~\cite{xing2022understanding} passes the attention map through an additional 1D convolutional layer to adjust its weights. The CTR-GCN~\cite{chen2021channel} refines the spatial attention mechanism in channel dimension to learn different dynamical features in each channel. The HA-GCN~\cite{xing2022skeletal} proposes a hybrid attention mechanism that mixes product and subtract attention maps together to enrich the dynamic features of different input streams. In the temporal dimension, the ST-GCN, AGCN, and PGCN process features by a single 2D convolutional kernel, while HA-GCN and CTR-GCN employ multi-scale temporal convolutional kernel introduced in the work~\cite{liu2020disentangling}.

The encoder is formed by concatenating $10$ aforementioned basic spatial-temporal graph convolutional blocks with different channel sizes.

\subsubsection{Decoder}

In order to upsample the condensed features back to the original time scale and predict action labels for each frame, we propose a novel temporal feature fusion decoder. As shown in Fig.~\ref{fig:decoder}, the decoder feature map is subscribed separately by two blocks, namely temporal feature extractor and feature fusion. The temporal feature extractor further compresses and extracts temporal features through three serial linear bottleneck layers~\cite{sandler2018mobilenetv2} and four parallel temporal pyramid average pooling~\cite{xing2022understanding} blocks. The four averaged feature outputs are concatenated, passed through a 2D convolutional kernel, and merged with a temporal residual connection. In the feature fusion block, the condensed feature is upsampled to the original time scale through one interpolate module. A depth-wise separable convolutional (\textit{DS Conv}) \cite{chollet2017xception} and 2D convolutional layers are employed to process the interpolated features. In the residual branch, the encoded condensed features are firstly processed by a 2D convolutional kernel and interpolated to the same size as in the main branch. The residual connection brings original high-dimensional features into the mainstream and fuses them with the low-dimensional features, which contributes to the performance accuracy. 

Furthermore, since the structure has good compatibility, we switch the proposed decoder with two existing upsampling methods, i.e., \textit{Fast-FCN}~\cite{wu2019fastfcn} and \textit{Temporal-Pyramid-Pooling} (TPP)~\cite{xing2022understanding}. The \textit{Fast-FCN} was originally introduced to solve the task of image segmentation and achieved promising results by jointly upsampling three processed feature maps from different depths of the encoder. To make the model suitable for action segmentation tasks, we convert the joint upsampling module to upsampling only on the time axis. The TPP is another recent decoder, which added four parallel \textit{temporal pyramid pooling} modules after a joint upsampling block. By utilizing dilated convolutional kernels with different scales, the TPP has a wide range of respective fields and achieves promising performance in action segmentation.

In the classifier, the fused features are first compressed in the spatial dimension by two depth-wise separable kernels and then mapped to the class space in the channel dimension.

\subsection{Distance-aware feature space by spectral normalized residual connection}

The residual connections improve the prediction performance by collapsing the features. However, in doing so, the distance in representation space is blurred, and the ability to detect out-of-distribution is further impaired. Hence, we propose a Spectral Normalized Residual connection to replace the traditional residual connection in the graph convolutional models, in which the mainstream has cascaded layers.

\textbf{Proposition:} Restricting the upper \textit{Lipschitz} bound in residual connections is essential to preserve feature space distances. The proof is in the appendix.

Considering a traditional residual connection using one convolutional kernel, where $r(\bm{x}) = {\phi}(\bm{Wx}+\bm{b})$, ${\phi}$, $\bm{M}$ and $\bm{b}$ are activation function, weight matrix, and bias respectively. We apply the spectral normalization on the weight matrix as follows:
\begin{equation}
    \bm{W}_{sn} = \begin{cases} 
    \bm{W}\cdot c/\lambda \ \ \ \; \  &c<\lambda\\
    \bm{W}\ \ \ \  \ \  \ \ \; \  \ &c\ge \lambda
    
    \end{cases}
\end{equation}
where $c > 0$ is a coefficient to adjust the norm bound of spectral normalization, and $\lambda$ is the spectral norm, i.e. the largest singular value of the weight matrix $\bm{W}$~\cite{behrmann2019invertible}. In doing so, we control the \textit{Lipschitz} upper bound of the residual connection by adjusting the hyperparameter $c$, since:
\begin{equation}
\begin{split}
        ||\sigma(\bm{W}_{sn}\bm{x}+\bm{b})||_{lip} \leq ||\bm{W}_{sn}\bm{x}+\bm{b}||_{lip} \\ 
        \leq ||\bm{W}_{sn}\bm{x} ||_{lip} \leq ||\bm{W}_{sn}||_{sn} \leq c
\end{split}
\end{equation}
where $||\cdot||_{lip}$ means \textit{Lipschitz} norm, e.g., $||\bm{W}_{sn}\bm{x} ||_{lip} = ||\bm{W}_{sn}\bm{x}_2 - \bm{W}_{sn}\bm{x}_1||/|| \bm{x}_2-\bm{x}_1||$, and $||\cdot||_{sn}$ represents the spectral norm.

\subsection{Feature space distance measurement using Gaussian process}
\begin{figure}[t]
  \centering
    \includegraphics[width=0.8\linewidth]{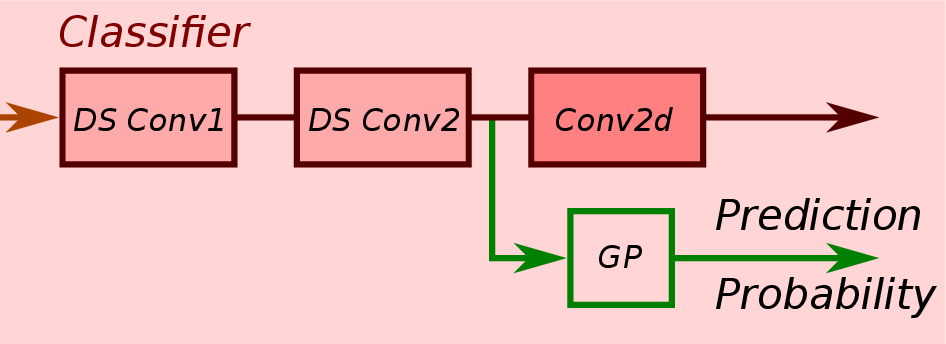}
  \caption{The Gaussian Process (GP) kernel collects high dimensional features from the network before the predictor and gives predictions with probabilities.}
  \label{fig:gp}
\end{figure}

By implementing the aforementioned model, we get a distance-aware feature space. We collect the high dimensional feature of all known data (trainset) output by the second separable kernel in the classier as shown in Fig~\ref{fig:gp}, and fit a multivariate normal distribution per class to quantify the prediction distance in the feature space, as follows:
\begin{equation}
    \bm{F}\sim N(\bm{\mu}, \bm{\Sigma})\text{,} \ \ \ \bm{\mu}\in \mathbb{R} ^{n\times c}, \ \bm{\Sigma}\in \mathbb{R} ^{n\times c\times c}
\end{equation}
where $c$ is the channel dimension of feature map $\mathbf{F}$, $n$ is number of action categories, $\bm{\mu}$ and $\bm{\Sigma}$ are the mean and covariance matrices, respectively. 

In the evaluation phase, we calculate the marginal likelihood of the unknown feature representation $\bm{f}^'$ under the prior density $\bm{F}$ per class: 
\begin{equation}
    p(\bm{f}^')_i = \sum_{j}^{c}p(f^'_j|f_{i,j})p(f_{i,j}), \ \ i \in [0, n-1]
\end{equation}
where $p$ is the Gaussian log probability, $c$ is number of channels,  $f^'_j$ and $f_{i,j}$ are the scalar elements of $\bm{f}'$ and $\bm{F}$, respectively. Since the log probability does not have the normalization ability like the \textit{softmax} function, the predicted label is selected by the one with the largest log probability and greater than a threshold. In doing so, the certainty of prediction is directly demonstrated by the log probability, and the feature space distance is transformed into the log probability space distance. 

In comparison, we utilize several existing measuring modules: the exponentiated distance~\cite{duq} and Laplace-approximated neural Gaussian process~\cite{sngp}


\section{Evaluations and results}
\label{sec:4}

\begin{table*}[t]
    \caption{Ablation study of introduced modules in the decoder on the Bimanual Actions dataset~\cite{datasetKIT}.}
	\label{tab:ablation_study_tf}
	\centering
	\resizebox{0.7\linewidth}{!}{%
		\begin{tabular}{>{\centering}m{0.5cm} >{\centering}m{0.5cm} >{\centering}m{0.5cm} >{\centering}m{0.5cm} >{\centering}m{0.5cm} >{\centering}m{0.5cm} >{}c >{}c >{}c >{}c >{}c}
	    \toprule
		\multicolumn{2}{ >{}l}{\textbf{GFE}$^a$} & \multicolumn{2}{ >{}l}{\textbf{TFF}} & \multicolumn{2}{ >{}l}{\textbf{Cl}} & \multicolumn{5}{ >{}l}{\textbf{Evaluation Metrics -  Bimanual Actions dataset}$^b$ ($\%$) } \\
		\cmidrule(lr){1-2} \cmidrule(lr){3-4} \cmidrule(lr){5-6} \cmidrule(lr){7-11}
		w/o & w & w/o & w & w/o & w & Top 1 $\uparrow$ & F1 macro $\uparrow$ & F1@10 $\uparrow$ & F1@25 $\uparrow$ & F1@50 $\uparrow$ 

        \\ 
        \cmidrule(lr){1-11}	
		  $\times$ & & $\times$ & & $\times$ & &
            $78.69$ & $79.23$ &
            $-$ &$-$& $-$ 
        \\ 
            & $\times$  & $\times$ & & $\times$ & &
            $79.85$ & $80.28$ & 
            $89.73$ & ${88.11}$ & ${{79.36}}$
        \\

		  $\times$ & & & $\times$ & $\times$ & &
            ${80.15}$ & ${80.08}$ & 
            ${88.66}$ & ${86.23}$ & $77.29$
        \\ 
            & $\times$ & & $\times$ & $\times$ & &
            $\underline{81.80}$ & $\underline{82.56}$ &
            $\underline{{90.49}}$ & $\underline{{87.99}}$ & ${\underline{78.92}}$ 
        \\ 
        
		  $\times$ & & $\times$ & & & $\times$ &
            ${82.97}$ & ${83.73}$ & 
            ${88.52}$ & ${86.63}$ & $77.23$
        \\ 
            & $\times$ & $\times$ & & & $\times$ &
            ${{86.11}}$ & ${{86.61}}$ &
            ${{90.48}}$ & ${{89.09}}$ & ${{80.43}}$ 
        \\  
		  $\times$ & & & $\times$ & & $\times$ &
            ${84.21}$ & ${85.04}$ & 
            ${87.83}$ & ${85.48}$ & ${{76.66}}$ 
        \\	

            & $\times$ & & $\times$ & & $\times$ &
            $\underline{\bm{89.06}}$ & $\underline{\bm{89.24}}$ &
            $\underline{\bm{93.82}}$ & $\underline{\bm{92.27}}$ & ${\underline{\bm{85.34}}}$ 
        \\
        \bottomrule

		\end{tabular}
	}
	\scriptsize
	\begin{tablenotes}
    	{\item[a]$^a$ The configurations are denoted as: ``GFE" = global feature extraction; ``TFF" = temporal feature fusion; ``CL" = classifier;``w" = with; ``w/o" = without. In setups without the temporal feature fusion module, the \textit{interpolate} function is employed for upsampling. In configurations lacking a \textit{classifier}, the spatial dimension is eliminated through averaging.
    	\item[b]$^b$ The experiments are conducted on the subject $1$ testset of the Bimanual Actions dataset~\cite{datasetKIT}. The best results across all configurations are in \textbf{bold}; The best results for each setup with or without the Classifier are \underline{underlined}.}
        \end{tablenotes}
\end{table*}

To evaluate the performance of the proposed UQ-TFGCN model, we experiment on two public challenging human-object interaction datasets: Bimanual Actions dataset \cite{datasetKIT} and IKEA
Assembly dataset \cite{datasetIKEA}.  A detailed ablation study is performed on the Bimanual Actions dataset \cite{datasetKIT} to examine the contributions of the proposed model components. Then, the final model is evaluated on both datasets and the results are compared with other state-of-the-art methods. 

All experiments are conducted on the PyTorch deep learning framework with a single NVIDIA-2080ti GPU. The stochastic gradient descent (SGD) with Nesterov momentum (0.9) is selected as the optimization strategy. A $16$ batch size is used in the training process, and a $1$ batch size is applied for testing each entire record. Cross-entropy is adopted as the loss function for the backpropagation. The weight decay is set to $0.0002$.

\subsection{Evaluation metrics}
Top 1 accuracy, F1-macro (weighted) score, and macro-recall are selected to evaluate the performance of recognition accuracy. F1@k is used to compare the performance of action segmentation, and k is set to $10\%, 25\%, 50\%$, same as in the work \cite{xing2022understanding}. \textit{The area under the receiver operating characteristic} (AUROC) and \textit{the area under the precision-recall curve} (AUPRC) metrics are chosen to measure the performance of \textit{out-of-distribution} (OOD) detection. Since the softmax output loses the true feature distance by normalization, we use Gaussian lop probability values of the multivariate model as the measurement score for AUROC and AUPRC. As recommended in the work~\cite{nixon2019measuring}, \textit{static calibration error} (SCE), \textit{adaptive calibration error} (ACE), and
\textit{thresholded adaptive calibration error} (TACE) are adopted to measure the calibration error, where the threshold is set as $0.01$. The popular \textit{expected calibration error} (ECE) is additionally compared, although it is designed for binary classification methods.

\subsection{Human-object interactions dataset}
\textbf{Bimanual Actions Dataset} (BimActs) \cite{datasetKIT} was built for human-object interaction detection in daily life environments, which has in total 540 recordings ($\sim$ $2.30$h). Framewise predictions of 6 subjects (3D skeletons) and 12 objects (3D bounding boxes) are provided. An evaluation benchmark is recommended by the authors: \textbf{leave-one-subject-out} cross-validation that contains records from one subject for validation and the rest subjects for training. To evaluate distance-aware performance, we create \textbf{Noisy BimActs} as noisy \textit{out-of-distribution} (OOD), which are with the same number of frames as the original data, but with $50\%$ impulse noise and empty data, respectively. In our experiments, we split each training record into small segments of 120 frames every 10 frames. In the test phase, the entire record of test data is fed into the model without any preprocessing. It should be noted that all noise is introduced at the node level. Thus, $50\%$ empty data indicates that half of the nodes (including skeleton joints and objects) are not detected.

\textbf{IKEA Assembly Dataset} (IKEA ASM) \cite{datasetIKEA} is another challenging and complex human-object interaction dataset, which contains $16,764$ assembly actions in $32$ categories ($\sim$ $35.27$h). 2D skeletons and 2D object bounding boxes are provided for each frame. The authors proposed a \textbf{cross-environment} benchmark, in which the test environments (117 scans) do not appear in the trainset (254 scans) and vice-versa. To evaluate the performance on unknown \textit{out-of-distribution} detection, we regard the testset of IKEA ASM as a completely \textit{out-of-distribution} (OOD) to the model trained on the BimActs dataset. The preprocessing of train data is similar to the BimActs dataset but with a segment length of $500$ frames per $100$ frame gap.

For both the BimActs~\cite{datasetKIT} and IKEA Assembly datasets~\cite{datasetIKEA}, The original 3D/2D skeleton and object position data are concatenated along the node dimension. It results in a tensor of shape $B \times C \times T \times V$, where $B$ denotes the batch size, $C$ represents the number of channels, $T$ corresponds to the number of frames, and $V$ indicates the number of nodes. Before being fed into the model, the data undergoes preprocessing that involves the application of controlled noise in rotation, translation, and scaling.
Specifically, rotational noise was introduced by perturbing the data within a range of $[-10^\circ, 10^\circ]$ around the z-axis. Translational adjustments were applied within a range of $[-0.2, 0.2]$ meters or pixels along the x and y axes, and scaling variations were implemented by applying a multiplicative factor between $[0.9, 1.1]$ on both the x and y axes. Note that all x, y, and z axes are in camera or image coordinates.

\begin{table*}[t]
    \caption{Comparison of encoder-decoder setups on the Bimanual Actions dataset~\cite{datasetKIT}$^a$.}
    \fontsize{6pt}{6pt}\selectfont
    \centering
    \label{tab:num_param}
    \resizebox{\linewidth}{!}
    {%
		\begin{tabular}{c c c c c c c c}
		\toprule
		Encoder & Decoder & Top 1 $\uparrow$ & F1 micro $\uparrow$ & F1@10 $\uparrow$ & F1@25 $\uparrow$ & F1@50 $\uparrow$ & $\#$Parameters \\ \cmidrule(lr){1-8}
			\multirow{3}{*}{\textit{ST-GCN}~\cite{yan2018spatial}} & \textit{Fast-FCN~\cite{wu2019fastfcn}} & $78.97$ & $82.05$ & $71.58$ & $68.34$ & $58.13$ & $4,503,842$ 
			\\ 
			& \textit{TPP~\cite{xing2022understanding}} & $78.88$ & $82.18$ & $83.42$ & $80.01$ & $69.40$ & $5,051,042$
			\\ 
			& \textit{TF} & $\underline{81.67}$ & $\underline{84.20}$ & $\underline{85.53}$ & $\underline{82.61}$ & $\underline{71.57}$ & $21,821,858$ \\
		    \multirow{3}{*}{\textit{AGCN}~\cite{shi2019two}} & \textit{Fast-FCN~\cite{wu2019fastfcn}} & $79.03$ & $81.98$ & $73.70$ & $69.99$ & $59.89$ & $4,874,370$ 
		    \\ 
		    & \textit{TPP~\cite{xing2022understanding}} & $84.13$ & $85.48$ & $85.93$ & $83.61$ & $74.56$ & $5,421,570$ 
		    \\ 
		    & \textit{TF} & $\underline{84.66}$ & $\underline{87.10}$ & $\underline{88.51}$ & $\underline{85.79}$ & $\underline{75.92}$ & $22,192,386$ \\ 
			\multirow{3}{*}{\textit{CTR-GCN}~\cite{chen2021channel}} & \textit{Fast-FCN~\cite{wu2019fastfcn}} & $79.52$ & $81.94$ & $64.97 $ & $61.83$ & $52.35$ & $2,811,296$ \\
			& \textit{TPP~\cite{xing2022understanding}} & $84.70$ & $87.33$ & $86.03$ & $84.02$ & $75.36$ & $3.358.496$ \\
			& \textit{TF} & $\underline{\bm{89.06}}$ & $\underline{\bm{89.24}}$ & $\underline{\bm{93.82}}$ & $\underline{\bm{92.27}}$ & $\underline{\bm{85.34}}$ & $20,129,312$
			\\
			\multirow{3}{*}{\textit{HA-GCN}~\cite{xing2022skeletal}} & \textit{Fast-FCN~\cite{wu2019fastfcn}} & $85.43$ & $87.30$ & $80.91$ & $78.61$ & $70.59$ & $2,821,568$ \\ 
			& \textit{TPP~\cite{xing2022understanding}} & $84.81$ & $87.08$ & $81.47$ & $78.83$ & $71.08$ & $3,368,768$ \\ 
			& \textit{TF} & $\underline{85.46}$ & $\underline{87.64}$ & $\underline{91.08}$ & $\underline{88.75}$ & $\underline{77.59}$ & $20,139,584$ \\
			\multirow{3}{*}{\textit{PGCN $^b$}~\cite{xing2022understanding}} & \textit{Fast-FCN~\cite{wu2019fastfcn}} & $80.13$ & $82.65$ & $70.06$ & $66.37$ & $56.37$ & $4,874,400$ \\
			& \textit{TPP~\cite{xing2022understanding}} & $84.86$ & $86.75$ & $88.58$ & $85.82$ & $76.40$ & $5,421,600$ \\
			& \textit{TF} & $\underline{86.44}$ & $\underline{88.55}$ & $\underline{89.58}$ & $\underline{86.94}$ & $\underline{76.96}$ & $20,129,712$ \\ 
                \bottomrule
			\end{tabular}
	}
	\scriptsize
	\begin{tablenotes}
	        \item[a]$^a$ The experiments are conducted on the subject $1$ testset of the Bimanual Actions dataset~\cite{datasetKIT}. The best results across all setups are in \textbf{bold}. The best results of the decoder setup are \underline{underlined}.
	        \item[b]$^b$ The encoder of PGCN~\cite{xing2022understanding}. 
	\end{tablenotes}
\end{table*}

\subsection{Ablation studies}
To analyze the contribution of each module in the proposed decoder, an ablation study is conducted on the Bimanual Actions dataset ~\cite{datasetKIT}. Distinct configurations are individually integrated into the decoder, which incorporates variations with or without \textit{global feature extraction}, \textit{temporal feature fusion}, and \textit{classifier} modules. The results of the ablation study are demonstrated in Table~\ref{tab:ablation_study_tf}. All three proposed modules contribute to the performance of action recognition and segmentation. The \textit{classifier} exerts the most significant influence on the performance. As the final layer, it plays a crucial role in mapping feature maps to predictions.  Thanks to the wide field of view, \textit{global feature extraction} yields notable enhancements in both action recognition and segmentation. However, relying solely on temporal feature fusion does not contribute significantly to enhancement.  In the absence of global features, it merely upsamples the condensed features of the encoder to the original time scale. The synergy of these two modules archives optimal performances across configurations, with or without the classifier.

\begin{table*}[t]
    \caption{Ablation study of the Spectral Normalized Residual connection and the Gaussian process on the Bimanual Actions dataset~\cite{datasetKIT}.}
	\label{tab:ablation_study}
	\centering
	\resizebox{\linewidth}{!}{%
		\begin{tabular}{>{\centering}m{0.5cm} >{\centering}m{0.5cm} >{\centering}m{0.5cm} >{\centering}m{0.5cm} >{\centering}m{0.5cm} c c c c c c c c}
	    \toprule
		\multicolumn{3}{l}{\textbf{Res}$^a$} & \multicolumn{2}{l}{\textbf{GP}} & \multicolumn{8}{l}{\textbf{Evaluation Metrics -  Bimanual Actions dataset}$^b$ ($\%$) } \\
		\cmidrule(lr){1-3} \cmidrule(lr){4-5} \cmidrule(lr){6-13}
		w/o & w & SN & w/o & w & Top 1 $\uparrow$ & F1 macro $\uparrow$ & F1@10 $\uparrow$ & F1@25 $\uparrow$ & F1@50 $\uparrow$ 

        \\ 
        \cmidrule(lr){1-10}	
		  $\times$ & & & $\times$ & &
            $73.69\pm1.81$ & $74.24\pm1.79$ &
            $86.02\pm1.01$ &$82.51\pm1.20$& $\underline{71.26}\pm1.58$ 
        \\ 
        
		$\times$ & & & & $\times$ & 
            $\underline{75.46}\pm{1.82}$ & $\underline{76.78}\pm1.79$ & 
            $\underline{86.17}\pm1.01$ & $\underline{82.61}\pm1.19$ & $69.67\pm1.57$
        \\ 
        
		& $\times$ & & $\times$ & & 
            ${89.34}\pm 0.57$ & ${89.52}\pm0.59$ & 
            $93.29\pm0.67$ & $92.17\pm0.75$  & $85.08\pm1.17$ 
        \\ 
        
		& $\times$ & & &$\times$ & 
            $\underline{\bm{89.34}}\pm0.57$ & $\underline{\bm{89.54}}\pm{0.59}$ & 
            $\underline{93.30}\pm{0.64}$ & $\underline{92.17}\pm{0.71}$ & $\underline{\bm{85.10}}\pm{1.14}$ 
        \\	
		
            &  & $\times$ & $\times$ & &
            $88.14\pm0.30$ & $88.66\pm{0.29}$ & 
            $92.73\pm{0.29}$ & ${91.72}\pm{0.36}$ & ${{84.51}}\pm0.53$
        \\ 
	
            & & $\times$ & & $\times$ & 
            $\underline{{88.44}}\pm0.29$ & $\underline{{88.89}}\pm0.29$ &
            $\underline{\bm{93.31}}\pm{0.29}$ & $\underline{\bm{92.18}}\pm{0.33}$ & ${\underline{84.76}}\pm{0.53}$ 
        \\ 
        \cmidrule(lr){1-13}	
        w/o & w & SN & w/o & w & ECE$\downarrow$ & TACE$\downarrow$ & ACE$\downarrow$ & SCE$\downarrow$ & AUROC $^{1} \uparrow$ & AUPRC$^1 \uparrow$  & AUROC $^{2} \uparrow$ & AUPRC$^2 \uparrow$ \\
        \cmidrule(lr){1-13}	
		  $\times$ & & & $\times$ & &
            $\underline{13.98}\pm1.33$ & $11.59\pm0.42$ & $90.56\pm0.09$ & $91.42\pm0.08$ & 
            $-$ & $-$ & $-$ & $-$
        \\ 
        
		$\times$ & & & & $\times$ & 
            $20.98\pm1.74$ & $\underline{2.49}\pm0.14$ & $\underline{81.61}\pm0.93$ & $\underline{82.43}\pm0.94$  &
            ${93.04\pm7.75}$ & ${90.76\pm9.05}$ & $41.11\pm7.89$& $44.61\pm5.31$
        \\ 
        
		& $\times$ & & $\times$ & & 
            $\underline{\bm{8.21}}\pm0.21$ & $2.15\pm0.12$ & $91.75\pm0.02$ & $92.62\pm0.02$ &
            $-$ & $-$  & $-$ & $-$  
        \\ 
        
		& $\times$ & & &$\times$ & 
            ${10.47}\pm{0.34}$ & $\underline{\bm{0.70}}\pm0.03$ & $\underline{\bm{30.98}}\pm0.36$ & $\underline{\bm{31.69}}\pm0.36$ & 
            $72.82\pm3.41$ & $68.08\pm3.75$ & $83.74\pm1.11$ & $84.37\pm1.26$
        \\	
		
            &  & $\times$ & $\times$ & &
            $\underline{{8.49}}\pm0.21$ & $2.59\pm0.08$ & $91.66\pm0.02$ & $92.57\pm0.01$ &
            $-$ & $-$ & ${-}$ & ${-}$
        \\ 
	
            & & $\times$ & & $\times$ & 
            ${11.24}\pm{0.19}$ & $\underline{{0.76}}\pm{0.02}$ & $\underline{{34.57}}\pm{0.32}$ & $\underline{{35.28}}\pm{0.33}$ &
            ${\bm{99.39}\pm0.97}$ & ${\bm{99.13}}\pm1.00$  & $\bm{90.19}\pm0.92$ & $\bm{92.07}\pm0.93$
        \\ 
        \bottomrule

		\end{tabular}
	}
	\scriptsize
	\begin{tablenotes}
    	\item[a]$^a$ The configurations are denoted as: ``Res" = Residual connections; ``SN" = spectral normalized; ``w" = with; ``w/o" = without; ``GP" = Gaussian process kernel. The configuration without Gaussian process kernel is using \textit{softmax} to output prediction.
    	\item[b]$^b$ The evaluation results are averaged over 10 seeds. The best results across all configurations are in \textbf{bold}; The best results for each setup with or without Gauss Process kernel are \underline{underlined}. A smaller standard deviation value indicates greater robustness in the model.
        The evaluation metrics \{AUROC$^1$, AUPRC$^1$\} and \{AUROC$^2$, AUPRC$^2$\} are using the \textbf{IKEA Assembly dataset} and \textbf{Noisy BimActs} as OOD respectively.
        \end{tablenotes}
\end{table*}

To obtain the best encoder-decoder combination, we adopt several popular existing GCNs (ST-GCN~\cite{yan2018spatial}, AGCN~\cite{shi2019two}, CTR-GCN~\cite{chen2021channel}, HA-GCN~\cite{xing2022skeletal}, PGCN~\cite{xing2022understanding}) as encoders, and combine them with Fast-FCN~\cite{wu2019fastfcn}, TPP~\cite{xing2022understanding} and the proposed temporal fusion (TF) decoders. As shown in Table~\ref{tab:num_param}, our TF decoder outperforms the other two decoders on all evaluation metrics across different configurations. Among them, the configuration of CTR-GCN~\cite{chen2021channel} plus TF achieves the highest performance in terms of accuracy, F1 macro, and F1@k. However, it is clear that the novel decoder dramatically increases the number of parameters, which leads to an increase in computation and a higher hardware cost in the application process. From this perspective, the CTR-GCN~\cite{chen2021channel} encoder has a promising performance in terms of efficiency, which has the least number of parameters among all encodes. The setup of CTR-GCN~\cite{chen2021channel} (encoder) and TF (decoder) forms the backbone of our Temporal Fusion Graph Convolutional Network~(TFGCN). 



\begin{figure}[t]
  \begin{center}
    \includegraphics[width=\linewidth]{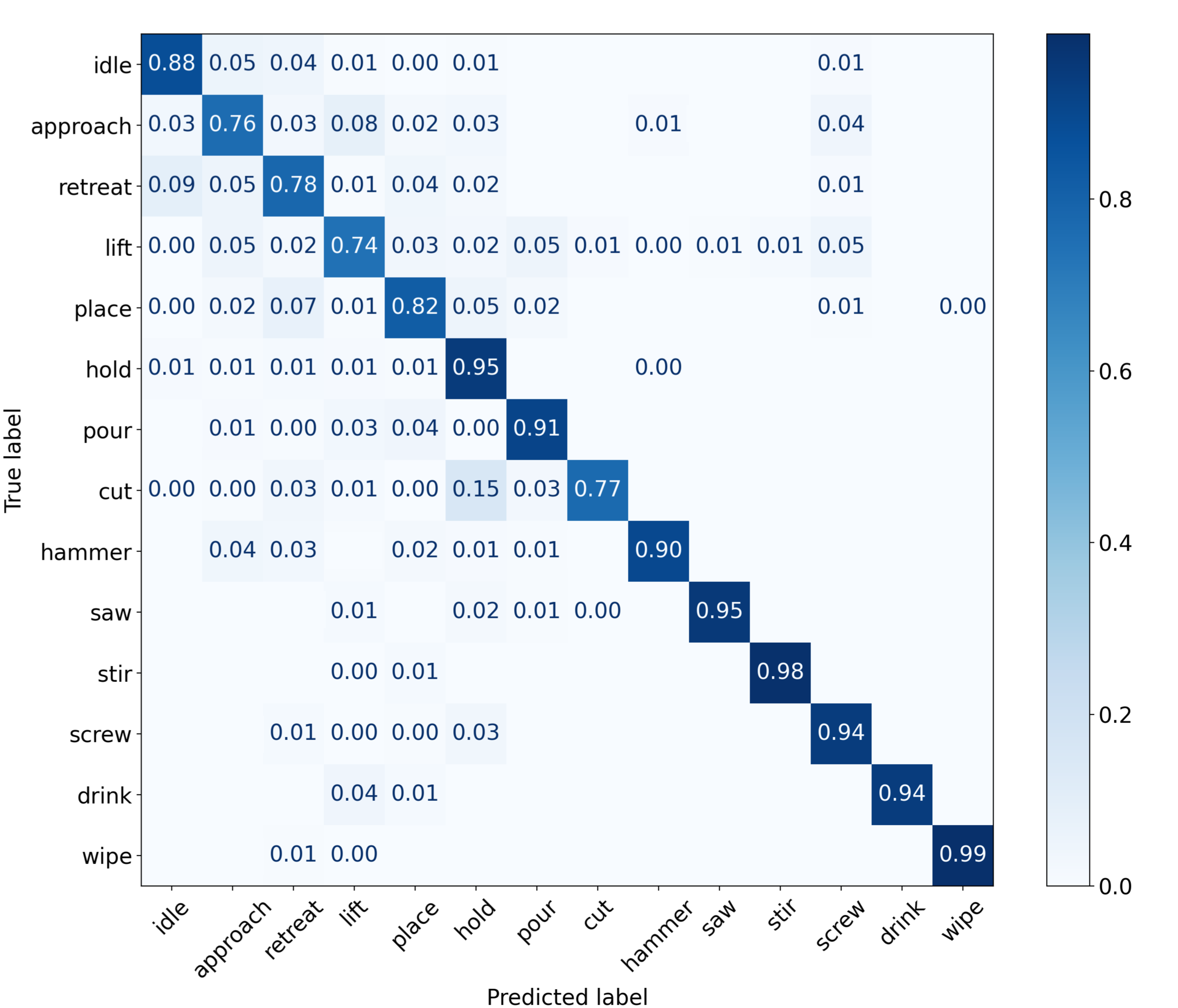}
  \end{center}
  \caption{Normalized confusion matrix for framewise prediction of accumulative classification correctness over the subject $1$ testset on the Bimanual Actions dataset~\cite{datasetKIT}}
  \label{fig:confusion}
\end{figure}

The normalized confusion matrices for the top prediction from the final model are presented in Fig~\ref{fig:confusion}. The principal challenge lies in accurately predicting actions such as \textit{approach}, \textit{lift}, \textit{retreat}, and \textit{place}, particularly when these actions involve interactions with diverse objects. This difficulty is amplified by the use of unstable object detection methods. Furthermore, the actions of \textit{approach} and \textit{retreat} typically occur instantaneously, occasionally taking place within $5$ frames. Another notable observation is the misclassification of “hold”, while the actual action is “cut”. The misclassification is caused by the utilization of "wrist" joints to represent "hands". This results in a limited resolution of motion detection and a potential misconception of the action as "hold”. These challenges could be mitigated through the implementation of stable object detection and human pose estimation methods, as well as additional hand pose estimation techniques. Nonetheless, rather than developing perfect algorithms for object detection, human pose estimation, and hand pose estimation, our primary concern revolves around the capability of the feature map to represent the input noise (input distance), and how to quantify the prediction uncertainty.

\begin{figure*}[t]
  \centering
    \begin{tabular}{@{}ccc@{}}
    \multicolumn{3}{c}{\includegraphics[width=0.8\linewidth]{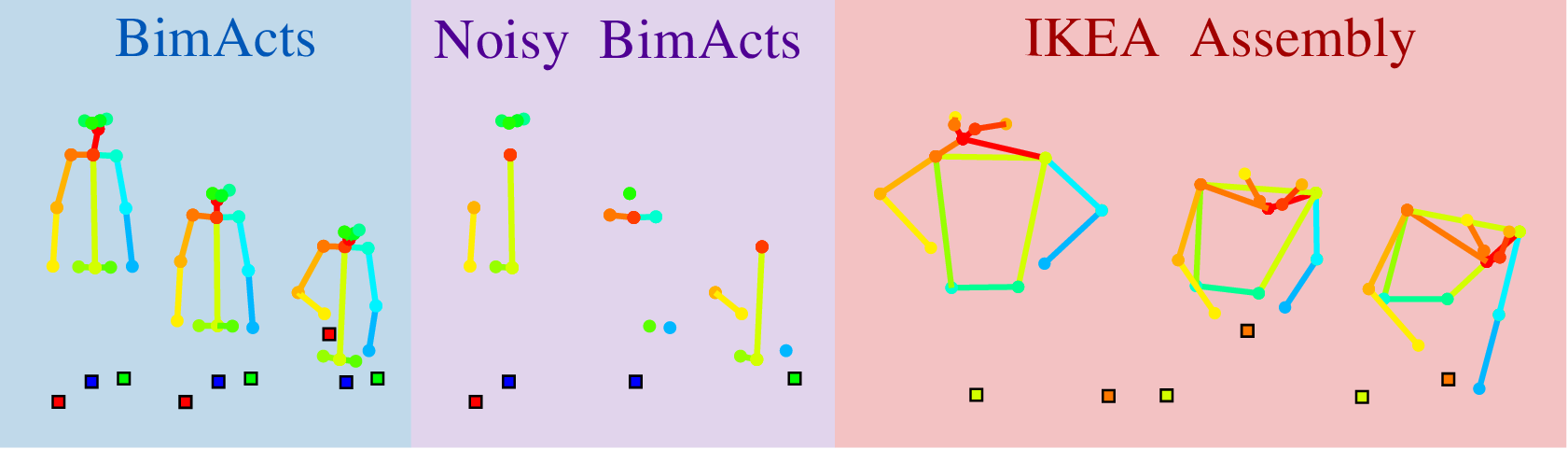}} \\
    \multicolumn{3}{c}{ \thead{(a) Left: Original Bimanual Actions dataset (BimActs)~\cite{datasetKIT}.\\ Middle: Noisy Bimanual Actions dataset. Right: IKEA Assembly dataset~\cite{datasetIKEA}}} \\
    \includegraphics[width=0.3\textwidth]{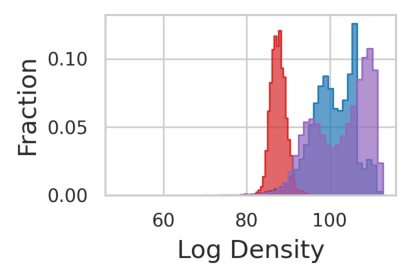} &
    \includegraphics[width=0.3\textwidth]{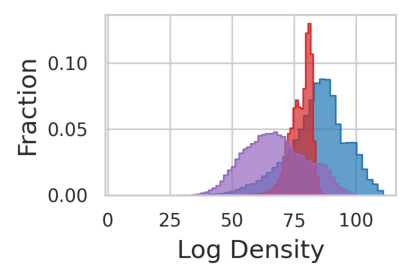} &
    \includegraphics[width=0.3\textwidth]{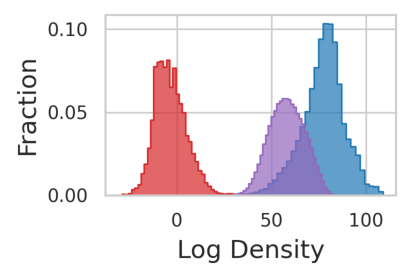} \\
    \multicolumn{3}{c}{\thead{(b) Gaussian log probability distribution. Left: Without residual connections.  \\ Middle: With normal residual connections. Right: With Spectral Normalized Residual connections}}\\
    \end{tabular}
  \caption{The log-probability distribution of feature space on the Bimanual Actions dataset \cite{datasetKIT}, noisy Bimanual Actions dataset, and the IKEA ASM \cite{datasetIKEA} datasets.}
  \label{fig:feature_density}
\end{figure*}

To analyze the influence of the residual connections on the feature space distance preserving and to evaluate the performance of the proposed Spectral Normalized Residual (SN-res) connection, we execute the experiments concerning three setups, namely without residual connections, with normal residual connections and with Spectral Normalized Residual connections. Table~\ref{tab:ablation_study} summarizes the ablation studies of the Spectral Normalized Residual connection, including Gaussian process (GP) configurations. By comparing the metrics produced by the Gaussian process and the \textit{softmax}, it is clear that the Gaussian process facilitates significantly fewer multi-class calibration errors, namely TACE, ACE, and SCE. In contrast, the configurations of \textit{softmax} have less Expected Calibration Error (ECE), because the ECE heavily relies on the overconfident probability of the predicted class from \textit{softmax}.

The results in Table~\ref{tab:ablation_study} confirm that the residual connection is of high significance for accurate predictions and accuracy-related performance, such as F1 macro and F1@k. Moreover, the results of the variance demonstrate that the residual connections increase the model stability. However, another remarkable observation is that adding residual connections leads to a smaller feature space distance (AUROC$^1$ and AUPRC$^1$) between the Bimanual Actions testset (\textit{in-distribution}) and the IKEA testset (\textit{out-of-distribution}), and a larger feature space distance between the noisy Bimanual Actions (Noisy BimActs) testset and the original testset. The observations confirm that the residual connections shift the \textit{bi-Lipschitz} bounds to a higher range, which leads to a higher distance to meaningful changes in the input manifold but lower sensitivity to \textit{out-of-distribution}. Note that, we consider the \textit{noisy BimActs} to have meaningful changes in the input manifold since it holds $50\%$ of the original data. 

The Spectral Normalized Residual connections set a maximum constraint on the upper \textit{Lipschitz} bound, which enhances the OOD detection while maintaining sensitivity to changes in the manifold. The results can also be observed in Fig~\ref{fig:feature_density}, where the feature space density is represented in the Gaussian log-probability space. However, it is clear that spectral normalization is a trade-off between preserving feature space distance and high accuracy. In Table~\ref{tab:ablation_study}, the accuracy performance of the spectral normalized model is $1\%$ lower than the model with normal residual connections, while its AUROC and AUPRC achieve the best result.

\begin{table*}[t]
    \caption{Comparison of coefficient values in spectral normalization function on the Bimanual Actions dataset~\cite{datasetKIT}.}
    \centering
    \label{tab:coefficience}
    \resizebox{0.7\linewidth}{!}
    {%
		\begin{tabular}{c c c c c c c c c}
		\toprule
		Coefficient $c$ & Top 1 $\uparrow$ & F1 micro $\uparrow$ & F1@10 $\uparrow$ & TACE $\downarrow$ & ACE $\downarrow$ & SCE $\downarrow$ & AUROC$^1\uparrow$ & AUPRC$^1\uparrow$ \\ \cmidrule(lr){1-9} 
			1 & $87.06$ & $87.75$ & $91.32$ & $0.86$ & $37.11$ & $37,83$ & $97.12$ & $96.58$ \\ 
		    2 & $88.24$ & $88.77$ & $92.63$ & $0.78$ & $34.86$ & $35.29$ & $\bm{99.94}$ & $\bm{99.88}$ \\ 
			3 & $88.44$ & $88.89$ & $93.31$ & $0.76$ & $34.57$ & $35.28$ & $99.39$ & $99.13$ \\ 
			4 & $88.68$ & $89.02$ & $93.59$ & $0.75$ & $32.37$ & $33.17$ & $89.81$ & $87.21$ \\ 
			5 & $\bm{89.27}$ & $\bm{89.60}$ & $\bm{94.37}$ & $\bm{0.71}$ & $\bm{32.20}$ & $\bm{32.91}$ & $85.26$ & $76.39$ \\ 
                \bottomrule
			\end{tabular}
	}
	\scriptsize
	\begin{tablenotes}
	    \item[] The experiments are conducted on the leave-subject-one-out testset of the Bimanual Actions dataset~\cite{datasetKIT}. The evaluation experiments are performed with \textit{Gaussian Process} and the results are averaged over 10 seeds. The best results across all setups are in \textbf{bold}.
	\end{tablenotes}
\end{table*}

\begin{table*}[t]
  \caption{Comparison of cross-validation results with state-of-the-art methods on the Bimanual Actions dataset~\cite{datasetKIT}$^a$}\label{tab:sota_bimacts}
  \centering
  \resizebox{0.7\linewidth}{!}{%
  \begin{tabular}{l c c c c c}
    \toprule
      \multirow{2}{*}{Model} & Accuracy$^a$ & F1 macro & \multicolumn{3}{c}{F1@ ($\%$)}  \\ \cmidrule(lr){4-6}
      &  ($\%$) &  ($\%$) & 10  & 25 & 50  \\
    \cmidrule(lr){1-6}
      Dreher et al. \cite{datasetKIT} & $63.0$ & $64.0$ & $40.6 \pm 7.2$ & $34.8 \pm 7.1$ & $22.2 \pm 5.7$ \\
      H2O+RGCN~\cite{lagamtzis2023exploiting} & 68.0 & 66.0 & $-$ & $-$ & $-$ \\
      Independent BiRNN \cite{morais2021learning} & $74.8$ & $76.7$ & $74.8 \pm 7.0$ & $72.0 \pm 7.0$ & $61.8 \pm 7.3$ \\
      Relational BiRNN \cite{morais2021learning} & $77.5$ & $80.3$ & $77.7 \pm 3.9$ & $75.0 \pm 4.2$ & $64.8 \pm 5.3$  \\
      2G-GCN~\cite{qiao2022geometric} & $-$ & $-$ & $85.0\pm 2.2$ & $82.0 \pm 2.6$ & $69.2 \pm 3.1$ \\
      PGCN~\cite{xing2022understanding} & ${86.8}$ & ${83.9}$ & ${88.5}\pm {\bm{1.1}}$ &${85.5} \pm {2.0}$ & ${77.0} \pm 3.4$\\
      \textbf{UQ-TFGCN} (Ours) & $\bm{88.4}$ & $\bm{88.6}$ &$\bm{93.7} \pm 1.2$ & $\bm{91.9} \pm{\bm{1.6}}$ & $\bm{85.4} \pm {\bm{2.9}}$  \\
    \bottomrule
  \end{tabular}}
  \scriptsize
  \begin{tablenotes}
        \item[a]$^a$ The models are cross validated on the leave-one-subject-out benchmark, the best results of each class are in \textbf{bold}. The F1 micro and macro results are averaged, 
        and F1@k are listed with mean and standard deviation. 
        A smaller standard deviation value indicates greater robustness in the model.
  \end{tablenotes}
\end{table*}

\begin{table*}[t]
  \caption{Action recognition and segmentation results in terms of top-1 accuracy, macro-recall, and F1@k on the IKEA Assembly dataset \cite{datasetIKEA}}\label{tab:sota_ikea}
  \centering
  \resizebox{0.7\linewidth}{!}{%
  \begin{tabular}{l c c c c c}
    \toprule
      \multirow{2}{*}{Model} & Accuracy & Macro-recall & \multicolumn{3}{c}{F1@ ($\%$)}  \\ \cmidrule(lr){4-6}
      &  ($\%$) &  ($\%$) & 10  & 25 & 50  \\
    \cmidrule(lr){1-6}
    HCN~\cite{li2018co} & $39.15$ & $28.18$ & - & - & -  \\
    ST-GCN~\cite{yan2018spatial} &  $43.40$ & $26.54$ & - & - & -  \\
    multiview+HCN~\cite{datasetIKEA} &  $64.25$ & ${46.33}$ & - & - & -  \\
    ST-GCN+TPP~\cite{xing2022understanding} & $68.92$ & $25.63$ & $66.92$ & $59.66$ & $41.33$  \\
    AGCN+TPP~\cite{xing2022understanding} & $70.53$ & $27.79$ & $76.32$ & $69.85$ & $52.14$  \\
    MGAF~\cite{kim2021motion} & ${72.40}$ & ${49.10}$ & $-$ & $-$ & $-$\\
    CTR-GCN+TPP~\cite{xing2022understanding} & $78.70$ & $37.98$ & $78.84$ & $72.68$ & ${54.40}$   \\
    PGCN~\cite{xing2022understanding} & ${79.35}$ &${38.29}$ & ${81.53}$ & ${76.28}$ & ${58.07}$\\
    PIFL$^*$~\cite{yan2023progressive} & $\bm{84.60}$ & $\bm{62.00}$ & $-$ & $-$ & $-$\\
    \textbf{TFGCN} ({Ours}) & ${80.39}$ &${39.77}$ & $\bm{83.99}$ & $\bm{80.04}$ & $\bm{68.00}$\\
    \textbf{UQ-TFGCN} ({Ours}) & ${79.99}$ &${39.72}$ & ${82.11}$ & ${76.92}$ & ${64.81}$\\
    \bottomrule
  \end{tabular}}
  \scriptsize
  \begin{tablenotes}
        \item[a]$^*$ The model extracts appearance features by an I3D~\cite{carreira2017quo} model instead of using the provided object detection results from the dataset.
  \end{tablenotes}
\end{table*}

These observations motivate us to further analyze the influence of the coefficient value of spectral normalization function in the quantitative analysis. By comparing the results in Table~\ref{tab:coefficience}, it can be seen that a higher coefficient value facilitates better performance in terms of recognition and segmentation accuracy, but leads to a lower feature space distance when detecting \textit{out-of-distribution}, while lower coefficients do the opposite. When the coefficient is too small, such as $1$, the model converges to a local minimum, which greatly reduces the accuracy and also affects the distance preserving in the feature space. To balance the performance of accuracy and \textit{out-of-distribution} detection, we select $3$ as the optimal coefficient. This trade-off aligns with our hypothesis that spectral normalization preserves isometric properties in the model, thereby compromising the nonlinear mapping capabilities of the learning method. We contend that this compromise is justifiable for the sake of achieving a more comprehensive understanding of motion. The fine-tuning of the coefficient is task-dependent. For example, image classification and segmentation can employ similar values, as both tasks involve mapping the pixel range $[0, 255]$ to the number of classes. In contrast, the range of human motion is inherently unknown before it occurs and varies across different motion tasks.




\subsection{Comparison with state-of-the-art}

\begin{table*}[t]
	\caption{Uncertainty quantification performance on the Bimanual Actions dataset$^a$~\cite{datasetKIT}.}
	\label{tab:uncertainty_noise}
	\centering
	\resizebox{1.0\linewidth}{!}{%
	\begin{tabular}{l c c c c c c c}
	\toprule
		Method	& Accuracy $\uparrow$ & F1 macro $\uparrow$ & F1@10 $\uparrow$ & F1@25 $\uparrow$ & F1@50 $\uparrow$ & {$\#$Parameters}
  \\ 
            \cmidrule(lr){1-7}
		MC-Dropout & $88.32\pm {\bm{0.21}}$ & $88.81\pm0.20$ & $\bm{93.87}\pm{\bm{0.20}}$ & $\bm{92.45}\pm{\bm{0.23}}$ & $84.45\pm{\bm{0.38}}$ & {$-$} \\ 
		Ensemble & $88.09\pm0.69$ & $88.56\pm0.67$ & $92.96\pm0.68$ & $91.65\pm0.80$ & $83.65\pm1.20$ & {$-$}\\ 
            DUQ$^b$~\cite{duq} & $83.55\pm1.06$ & $83.91\pm1.06$ & $91.22\pm0.66$ & $89.04\pm0.86$ & $81.12\pm1.3$ & {$20,264,026$}\\ 
		SNGP$^c$~\cite{sngp} & $87.63\pm{\bm{0.21}}$ &$88.27\pm{\bm{0.16}}$ & $91.64\pm0.35$ & $90.17\pm0.43$ & $83.03\pm0.44$ & { $21,207,998$}\\ 
		\textbf{UQ-TFGCN}& $\bm{88.44}\pm0.29$ & $\bm{88.89}\pm0.29$ & $93.31\pm{0.29}$ & $92.18\pm{0.33}$ & $\bm{84.76}\pm{0.53}$ & {$20,261,282$}\\ 
        
        \cmidrule(lr){1-8}
        Method	& TACE$\downarrow$ & ACE$\downarrow$ & SCE$\downarrow$ & AUROC $^{1} \uparrow$ & AUPRC$^1 \uparrow$  & AUROC $^{2} \uparrow$ & AUPRC$^2 \uparrow$ \\
        \cmidrule(lr){1-8}
        MC-Dropout & $0.77\pm{\bm{0.01}}$  & $34.12\pm0.23$ & $34.83\pm0.23$ & $\bm{99.95}\pm{\bm{0.02}}$ & $\bm{99.89}\pm{\bm{0.02}}$ & $78.00\pm{\bm{0.09}}$ & $81.38\pm{\bm{0.08}}$ \\
        Ensemble & $0.79\pm0.05$ & $\bm{14.97}\pm0.55$ & $\bm{15.77}\pm0.56$ & $99.81\pm0.22$ & $99.68\pm0.33$ & $86.29\pm2.61$ & $88.76\pm2.17$ \\
        DUQ$^b$~\cite{duq} & $90.27\pm0.15$ & $90.27\pm0.15$ & $90.57\pm0.15$ & $83.58\pm3.98$ & $80.96\pm5.56$ & $52.86\pm7.43$ & $57.00\pm8.98$ \\
        SNGP$^c$~\cite{sngp} & $90.78\pm0.03$ & $90.78\pm{\bm{0.03}}$ & $91.13\pm{\bm{0.03}}$ & $93.16\pm3.44$ & $84.67\pm7.91$ & $79.17\pm1.89$ & $78.03\pm1.55$ \\
        \textbf{UQ-TFGCN} & $\bm{0.76}\pm{0.02}$ & $34.57\pm{0.32}$ & $35.28\pm{0.33}$ & $99.39\pm0.97$ & $99.13\pm1.00$  & $\bm{90.19}\pm0.92$ & $\bm{92.07}\pm0.93$  \\
        \bottomrule
	\end{tabular}}
	\scriptsize
	\begin{tablenotes}
		\item[a]$^a$ The evaluation results are based on the Bimanual Actions dataset leave-subject-one-out testset and are averaged over 10 seeds. A smaller standard deviation value indicates greater robustness in the model. The best results across all configurations are in \textbf{bold}. The evaluation metrics \{AUROC$^1$, AUPRC$^1$\} and \{AUROC$^2$, AUPRC$^2$\} are using the \textbf{IKEA Assembly dataset} and \textbf{Noisy BimActs} as OOD respectively.
            \item[b]$^b$ The \textit{Radial Basis Funtion} (RBF) kernel~\cite{duq} is implemented to measure the distance to class centroids.
            \item[c]$^c$ The \textit{Laplace-approximated Neural Gaussian Process}~\cite{sngp} is utilized instead of our Gaussian Process.
	\end{tablenotes}
\end{table*}

\begin{figure*}[t]
  \centering
    \begin{tabular}{@{}ccc@{}}
    \includegraphics[width=0.3\textwidth]{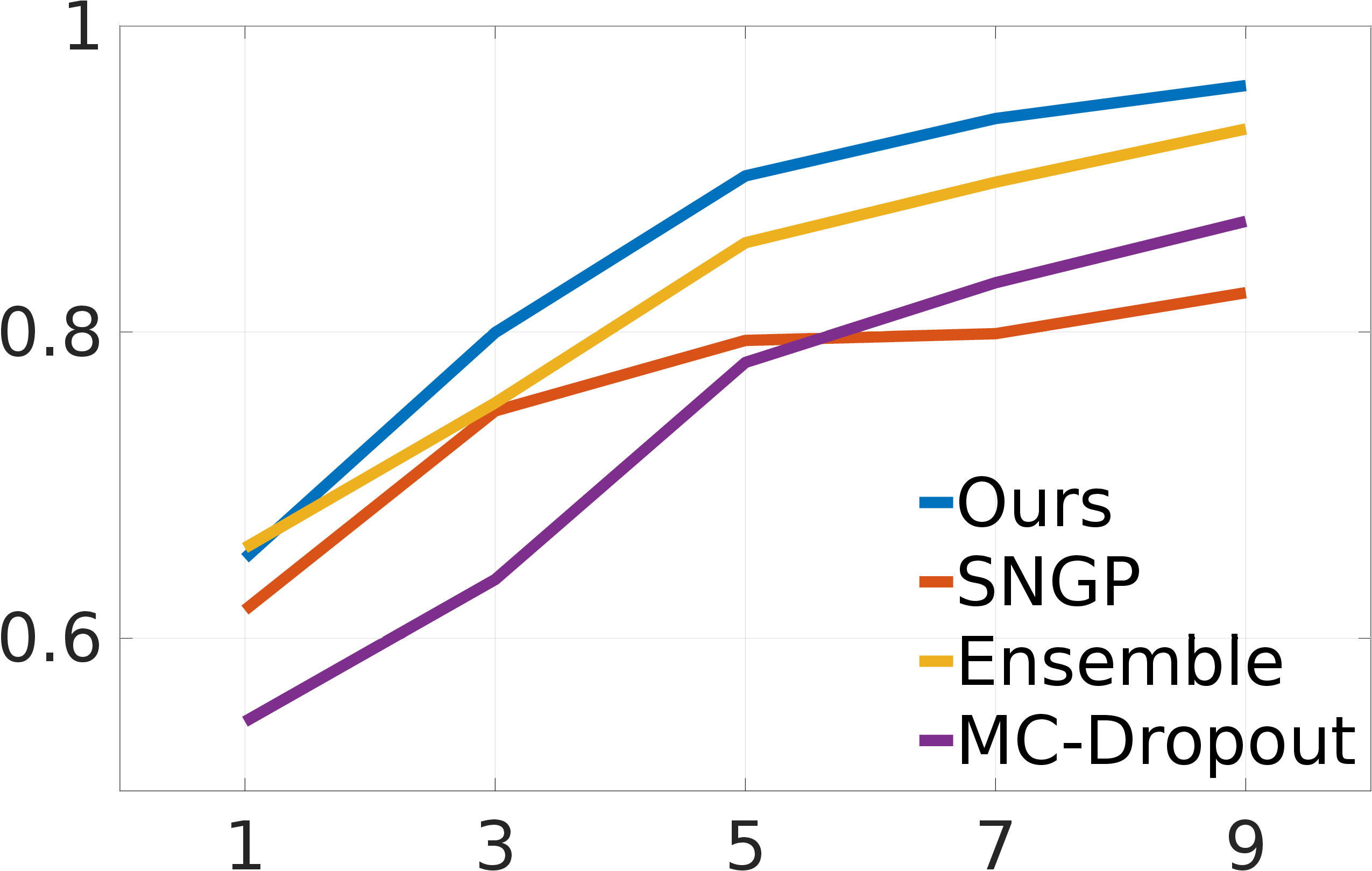} &
    \includegraphics[width=0.3\textwidth]{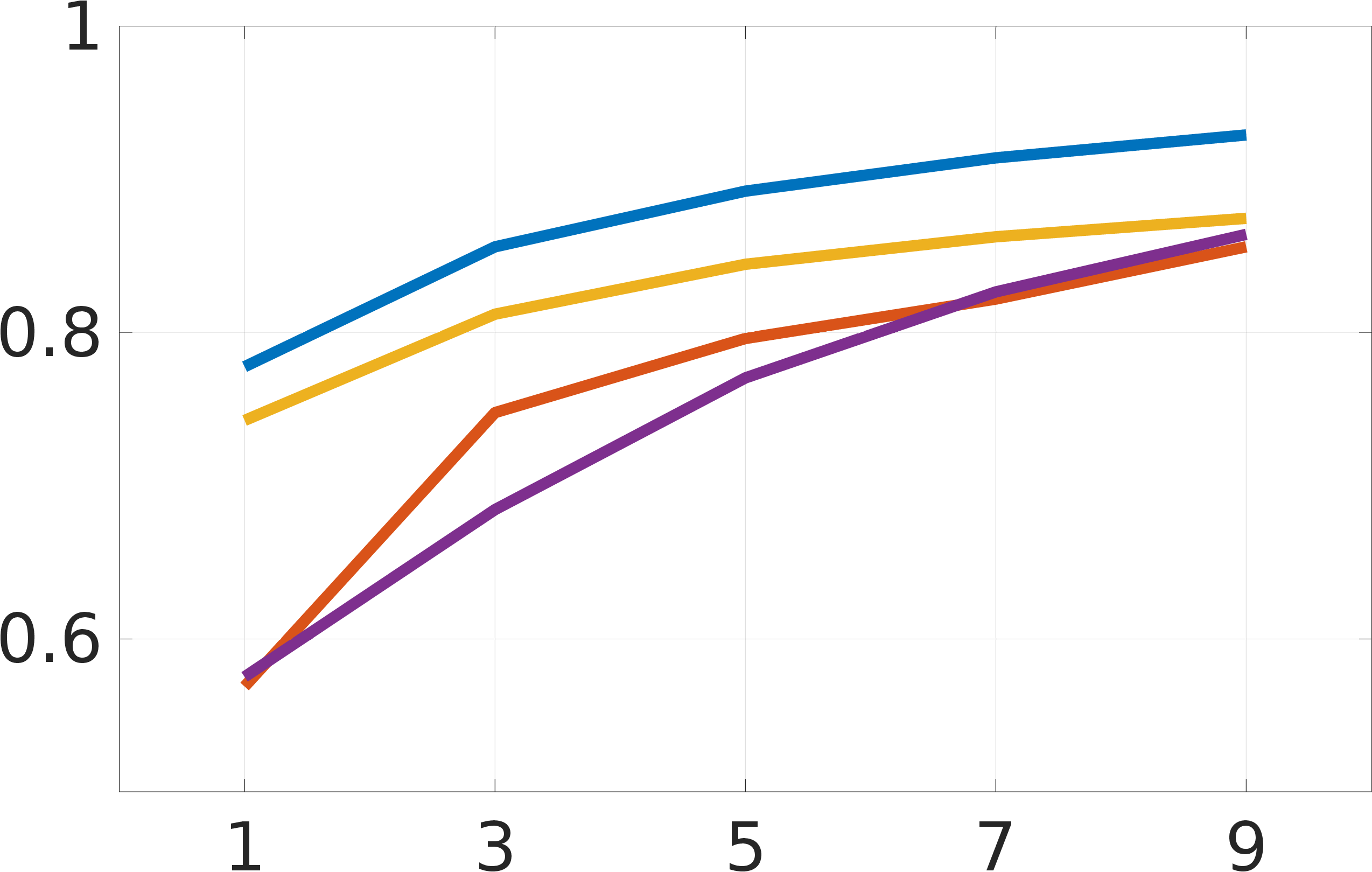} &
    \includegraphics[width=0.3\textwidth]{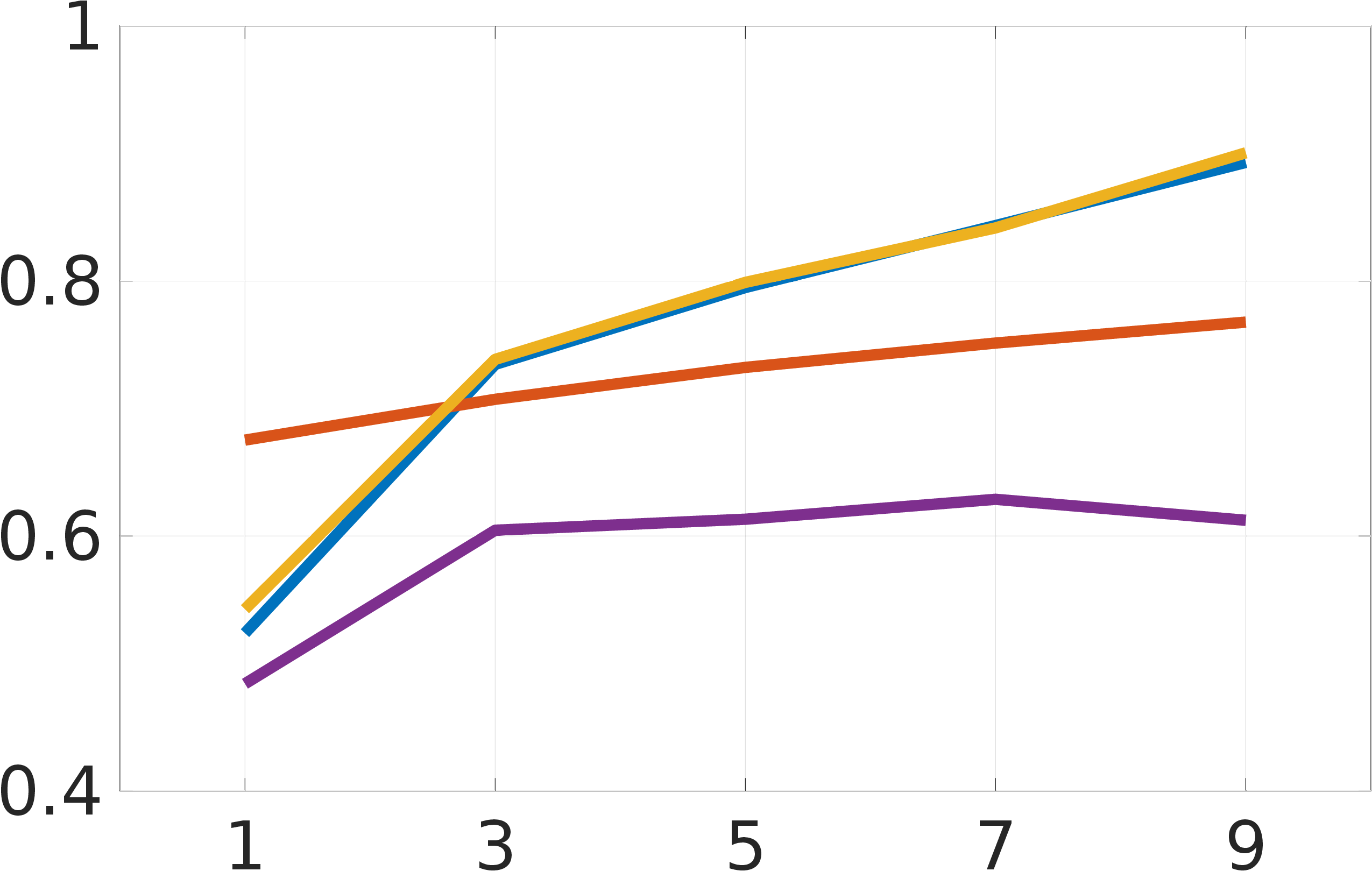} \\
    (a) Impulse noise & (b) Gaussian noise & (c) Poisson noise
    \end{tabular}
  \caption{Comparison of AUROC with increasing noise intensity in the Bimanual Actions Dataset \cite{datasetKIT} over different uncertainty quantification methods: Ensemble, MC-Dropout, SNGP, and ours.}\label{fig:noise}
\end{figure*}

To demonstrate the efficiency and robustness, we compare the performance of the proposed model with other popular existing methods on two challenging dataset in the field of Human-Object-Interaction recognition and segmentation, i.e., BimActs \cite{datasetKIT} and IKEA Assembly \cite{datasetIKEA} datasets. 
First, the quantitative experiment of action recognition and segmentation is conducted on the BimActs~\cite{datasetKIT} and IKEA Assembly~\cite{datasetIKEA} datasets, and the results are respectively listed in Table~\ref{tab:sota_bimacts} and Table~\ref{tab:sota_ikea}.

From Table~\ref{tab:sota_bimacts}, it is easy to see that our model achieves the best performance across all of the metrics on the Bimanual Actions dataset~\cite{datasetKIT}. Especially, the proposed {Uncertainty Quantified Temporal Fused Graph Convolutional Network} (UQ-TFGCN) significantly improves the performance in terms of average F1@k score (by $6.2\%$, $6.4\%$ and $8.4\%$) compared to the PGCN, which confirms its efficiency for action recognition and segmentation. 
Furthermore, the standard deviation in F1@k for our results is the smallest, indicating that the model exhibits high robustness.

The experimental results in Table~\ref{tab:sota_ikea} demonstrate that the proposed TFGCN outperforms the state-of-the-art methods in terms of f1@k segmentation score. 
While the PIFL~\cite{yan2023progressive} achieves superior performance on the IKEA Assembly dataset, it is unfair to directly compare it with other methods. This discrepancy arises from the fact that PIFL extracts appearance features using an I3D model, as opposed to utilizing the provided object detection results from the dataset. On the other hand, this underscores the significance of comprehensive instance information in the context of action determination. Another remarkable observation is that replacing the normal residual connection with Spectral Normalized Residual connections in UQ-TFGCN leads to a drop in accuracy and f1 score performance, which reaffirms the disruption of {spectral normalization}. The advantage of Spectral Normalized Residual connections cannot be demonstrated in these two experiments, because most existing recognition and segmentation models ignore the measure of feature distance.

\begin{figure*}[t]
  \centering
  \includegraphics[width=0.99\linewidth]{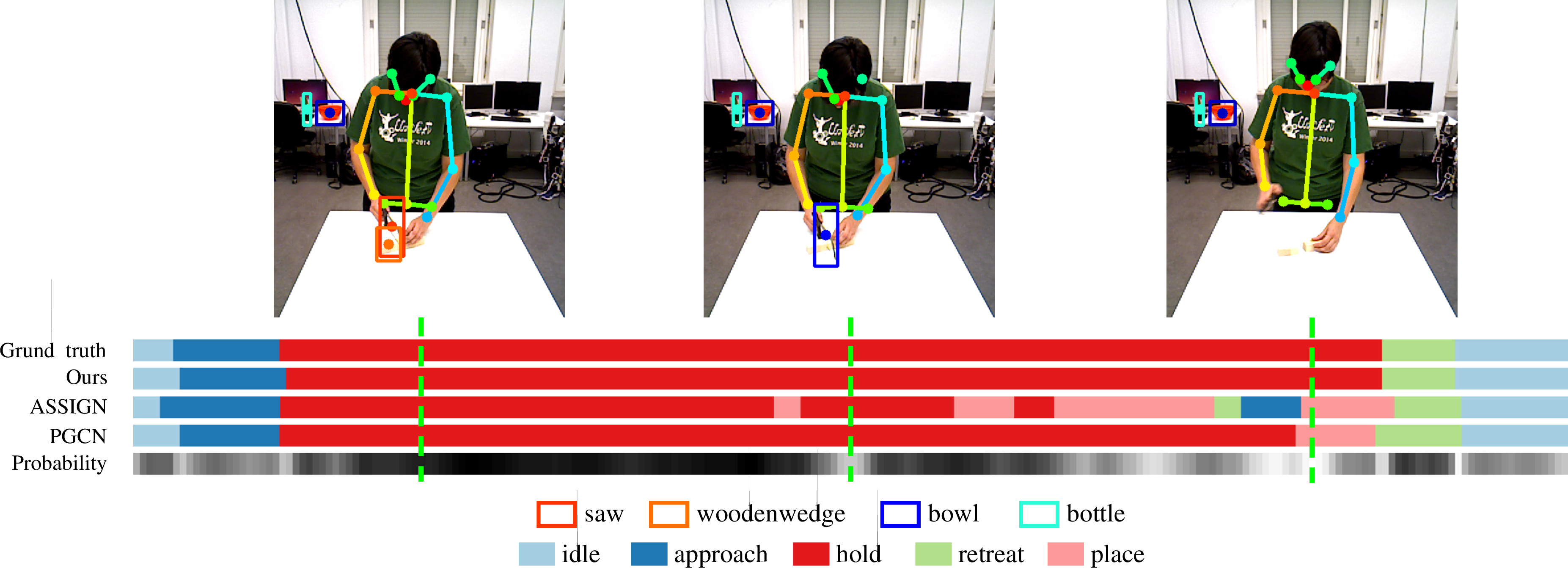} 
  \caption{Qualitative results for action recognition and segmentation of the left hand in an example of \textit{sawing} a wooden wedge from the Bimanual Actions dataset~\cite{datasetKIT}. Distinct actions are distinguished using various colors. The probability, generated by our UQ-TFGCN model, is visually represented in a grayscale bar, where brighter grayscale values correspond to lower predicted probabilities.}  \label{fig:qualitative1}
\end{figure*}

\begin{figure*}[t]
  \centering
  \includegraphics[width=0.99\linewidth]{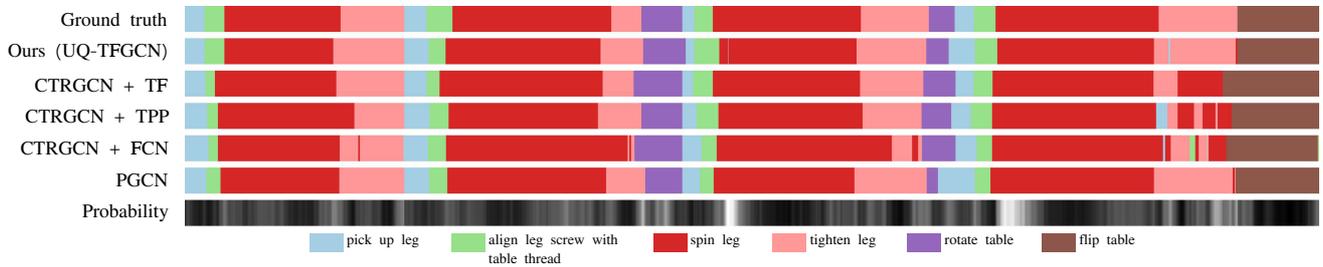} 
  \caption{Qualitative uncertainty estimation and activity segmentation results of \textit{assembly side table} example from the IKEA Assembly dataset~\cite{datasetIKEA}. Distinct actions are distinguished using various colors. A brighter grayscale value indicates a lower predicted probability.}  
  \label{fig:qualitative2}
\end{figure*}

In order to evaluate the performance of distance-awareness in the feature space, we conduct the \textit{out-of-distribution} (OOD) detection experiment and compare the results with other popular existing deterministic network uncertainty quantification methods, i.e., SNGP~\cite{sngp} and DUQ~\cite{duq}. Note that SNGP~\cite{sngp} and DUQ~\cite{duq} are not originally designed for action recognition and segmentation, therefore, we implement their feature space measuring mechanism instead of GP in our model. Besides that, the \textit{MC-Dropout} and \textit{ensemble} methods are utilized as the baseline. In the \textit{MC-Dropout} method, the dropout ratio is set as $10\%$. In the \textit{ensemble} method, the final prediction is given by three ensembled models. As shown in Table~\ref{tab:uncertainty_noise}, the proposed method outperforms all other listed methods in terms of accuracy, AUROC and AUPRC in the OOD of the noisy BimActs dataset. Both SNGP and DUQ are more efficient than our model, since they either approximate the \textit{Gaussian Process} by a \textit{Laplace process} (SNGP~\cite{sngp}) or initially give a covariance scale (DUQ~\cite{duq}), while our model collects features from trainset and build a multivariate Gaussian model. From the results, the multivariate Gaussian model has a better performance than the other two efficient methods. For different action categories, the covariance values of the Gaussian model range from $0.001$ to $15.00$. However, the DUQ only uses the initial manually given covariance scale to predict the probability, which is the main reason for the poor performance. The efficiency of incorporating SN (spectral normalization) into the residual connection is exemplified by the minimal number of parameters in our model. Note that each model utilized in the MC-Dropout and Ensemble methods has the same number of parameters as our model. From the perspective of standard deviation values, traditional MC-Dropout demonstrates a more robust performance compared to other methods. This discrepancy becomes more evident in OOD detection compared to DUQ~\cite{duq} and SNGP~\cite{sngp}, as indicated by AUROC and AUPRC metrics.
Another interesting observation is that randomly dropping features in \textit{MC-Dropout} leads to a poorer noisy OOD detection performance, which indicates that our model is sensitive to the known feature space. These observations motivate further research into the effects of different noise types and intensities.

Since noise is unavoidable in real scenarios, OOD detection on noisy datasets reveals some practical benefits of preserving distance in feature space. Three common noises in the real world are selected for comparison, namely \textit{impulse} noise, \textit{Gaussian} noise, and \textit{Poisson} noise. The corresponding results of different noise intensities are demonstrated in Fig~\ref{fig:noise}. Since each noise has its different intensity scale, we choose three increasing unit intensities for different noises. For the impulse noise, it set $10\%$ of testset to zeros by increasing one unit intensity. For the Gaussian noise, the increasing unit intensity is set as $0.1$ variance of testset. For the Poisson noise, it is $1000$~mm expected values and variance. As shown in Fig~\ref{fig:noise}, our model consistently maintains the highest AUROC in Gaussian and Impulse noise and achieves performance close to the MC-Dropout method in Poisson noise. The results demonstrate that our model is robust in preserving feature space distances under different noise conditions.

\subsection{Qualitative results}
We present the detailed outputs with the probability of our model and related methods on examples from Bimanual Action~\cite{datasetKIT} dataset. Fig~\ref{fig:qualitative1} shows an example of the left-hand actions in the \textit{sawing} activity. A remarkable achievement of our model is precise predictions of the end and the start of segments, which recognize
correctly the start and end frame index of \textit{retreat} in the example. It demonstrates that our model significantly improves the ability to prevent shift- and over-segmentation. Our model also achieves higher accuracy in action recognition compared to other methods. In the example, both ASSIGN~\cite{morais2021learning} and PGCN~\cite{xing2022understanding} predict a wrong \textit{place} action, and only our model successfully predicts a \textit{hold} action consistent with ground-truth. Another impressive achievement is that, in addition to the prediction of the action label, our model also outputs the predicted probability. As shown in the bottom of Fig~\ref{fig:qualitative1}, the probability represents the similarity between the current feature and known features from trainset, which makes the model's predictions more interpretable. Wrong predictions usually occur at the junction of two subactions, where the probability is low, because the features at this time are far away from the distribution centers of both actions. Predictions with low probabilities in a continuous action mean that the input has some noise, e.g., wrong object detection. In the example, we select three representative frames in a continuous hold action, i.e., predictions with high (black), medium (gray), and low (white) probabilities. At the position with high probability, the input has a precise location and correct label of objects, while at medium and low probability places, either objects are mislabeled or missing.

Another qualitative result of a complex \textit{assembly side table} task from the IKEA Assembly dataset~\cite{datasetIKEA} is demonstrated in Fig.~\ref{fig:qualitative2}. It can be seen that the proposed temporal fusion~(TF) decoder has a better performance in preventing shift- and over-segmentation compared with Fast-FCN~\cite{wu2019fastfcn} and \textit{temporal pyramid pooling} (TPP)~\cite{xing2022understanding} decoders.

\section{Conclusions}
\label{sec:5}


   We introduced a method to compute prediction uncertainty and preserve input distance in Human Action Recognition frameworks through a novel Uncertainty Quantified Temporal Fusion Graph Convolutional Network (UQ-TFGCN) for recognizing and segmenting Human-Object interaction sequences. The network consists of an attention-based graph convolutional network as an encoder and a temporal fusion module as a decoder. One novelty is the modification of the decoder to reduce errors in the up-sampling of features to prevent over-segmentation and to improve the detection of action class transitions. The new decoder leads to an increased number of parameters that increase the computational requirements. The resulting inference still allows online segmentation of actions on standard desktop hardware. Additionally, Spectral Normalized Residual connection helps to maintain meaningful isometric properties. At the same time, it harms the accuracy of the network. The accuracy level can be held at an acceptable level through parameter tuning trade-off between accuracy and feature space distance of the Spectral Normalized Residual.
   
   Experiments on public datasets demonstrate that the proposed temporal fusion decoder has significantly improved the model performance on human-object interaction recognition and segmentation, but it leads to a larger computation complexity as mentioned above. Experimental analysis on noise and out-of-distribution data detection proves that the proposed Spectral Normalized Residual connection is beneficial to preserve the input distance in the feature space, and contributes to estimating the uncertainty. The examination of parameter quantities provides compelling evidence supporting the efficiency of the proposed SN-res method. Results on public HOI datasets with two different data formats (2D and 3D) show that our model has a general capability that can be implemented on other structural-represented domains. 
   
   The study shows that the action recognition and segmentation capabilities of our model can be of high relevance for various use cases that require an understanding of human behavior, e.g., learning from demonstration and human-robot collaboration. Furthermore, the safety of collaborations between humans and robots can also be improved through an additional uncertainty output. Knowledge of the limitations of the acquired knowledge is only the first step. Active learning of new knowledge is the next natural step for an intelligent system. Therefore, we plan to extend the capabilities of our model to active learning based on novelty detection results.

\section*{Acknowledgment}
    \vspace*{-0.5\baselineskip}
	We gratefully acknowledge the funding of the Lighthouse Initiative Geriatronics by StMWi Bayern (Project X, grant no. 5140951).
	
    \vspace*{-0.5\baselineskip}
\section*{Appendix}
\label{sec:appendix}
Considering a graph convolutional layer $g(\bm{x})$ with residual connections: $g(\bm{x}) = r(\bm{x}) + m(\bm{x})$ where $m$ represents mainstream and composed of several hidden layers, $r$ is a residual branch and $\bm{x}$ is input. Assume that the upper \textit{Lipschitz} boundaries of main and residual streams are initially defined by the normalization and activation functions $\forall \bm{x}$, denoted as $\beta_m$ and $\beta_r$ respectively. The lower boundaries for both streams are defined as the extremum $0$. Note that in practice, the feature distance is unequal to $0$ due to non-zero weights and biases in the convolution kernels.
Simplify the processing functions to be $\bm{g}_{1,2}$, $\bm{r}_{1,2}$ and $\bm{m}_{1,2}$ where $1$ and $2$ mean with input $\bm{x}_1$ and $\bm{x}_2$ respectively, we get:
\begin{align}
     0 & \leq \frac{||\bm{m}_2 - \bm{m}_1 ||}{||\bm{x}_2 - \bm{x}_1 ||} \leq \beta_m, \\
    0 & \leq \frac{||\bm{r}_2 - \bm{r}_1 ||}{||\bm{x}_2 - \bm{x}_1 ||} \leq \beta_r.
\end{align}
where we simplify the symbol of \textit{Lipschitz} norm as $||\cdot ||$.

Additional residual connections shift the range to a higher value as follows:
\begin{equation}
\begin{split}
    ||\bm{r}_2- \bm{r}_1 ||=& ||\bm{g}_2- \bm{m}_2 - (\bm{g}_1 - \bm{m}_1)|| \\
     =& ||\bm{g}_2 - \bm{g}_1 + (\bm{m}_1 - \bm{m}_2)|| \\ 
     \leq & ||\bm{g}_1 - \bm{g}_2|| + ||\bm{m}_1 - \bm{m}_2|| \\
     \leq & ||\bm{g}_2 - \bm{g}_1|| + \beta_m||\bm{x}_2 - \bm{x}_1||,
\end{split}
\end{equation}
where the last line follows by the bound assumptions $\forall \bm{x}$, we get the lower bound range of $g(\bm{x})$:
\begin{align}
        ||\bm{r}_2 - \bm{r}_1 || - \beta_m||\bm{x}_2 - \bm{x}_1|| \leq ||\bm{g}_2 - \bm{g}_1|| \\
        \begin{gathered}
        -\beta_m \leq \frac{||\bm{r}_2 - \bm{r}_1 ||}{||\bm{x}_2 - \bm{x}_1||} - \beta_m\leq (\beta_r - \beta_m) \\ \leq \frac{||\bm{g}_2 - \bm{g}_1 ||}{||\bm{x}_2 - \bm{x}_1||}. 
        \end{gathered}
        \label{eq:residual_lower}
\end{align}

The upper bound can be easily obtained by:
\begin{equation}
\begin{split}
       \frac{||\bm{g}_2 - \bm{g}_1 ||}{||\bm{x}_2 - \bm{x}_1||} \leq & \frac{||\bm{r}_2 - \bm{r}_1 || + ||\bm{m}_2 - \bm{m}_1 ||}{||\bm{x}_2 - \bm{x}_1||} \\
       \leq & (\beta_r + \beta_m ).
\end{split}
\end{equation}

From Eq.~\ref{eq:residual_lower}, it can be seen that the upper bound of the residual stream shifts the feature space distance to a higher range when $\beta_r > \beta_m$. 
In fact, the mainstream produces fine feature maps through several cascaded layers, while the residual outputs coarse features, which means that the \textit{Lipschitz} upper bound of the feature space distance in the residual connection is larger than that of the mainstream, i.e., $\beta_r > \beta_m$. Note that when $\beta_r \leq \beta_m$, the feature space distance automatically satisfies the constraint, since $-\beta_m \leq \beta_r - \beta_m \leq 0$ and $0 \leq ||g_2 - g_1||$. Hence, constraining the \textit{Lipschitz} upper bound of residual connections is crucial to preserve distance in the representation space.

\bibliographystyle{SageH}
\bibliography{mybib.bib}

\end{document}